%% file: emnlp2022.tex
\pdfoutput=1
\documentclass[11pt]{article}

\usepackage[final]{EMNLP2022}

\usepackage{times}
\usepackage{latexsym}

\usepackage[T1]{fontenc}
\usepackage[utf8]{inputenc}

\usepackage{microtype}

\usepackage{inconsolata}
\usepackage{multirow}
\usepackage{bm}

\newcommand{\crtext}[1]{\textcolor{magenta}{#1}}

\usepackage[belowskip=2pt,aboveskip=0pt]{caption}
\usepackage{mathrsfs}

\usepackage{graphicx}
\usepackage{amsmath}
\usepackage{amssymb}
\usepackage{booktabs}

\usepackage{multirow}
\usepackage{appendix}
\usepackage{dutchcal}
\usepackage{bm}
\usepackage{booktabs,makecell,multirow}
\title{Prompting for Multimodal Hateful Meme Classification}

\author{
Rui Cao$^1$, Roy Ka-Wei Lee$^2$, Wen-Haw Chong$^1$, Jing Jiang$^1$
\\
$^1$Singapore Management University
\\
$^2$Singapore University of Design and Technology
\\
$^1$\{ruicao.2020@phdcs.,whchong.2013@phdis.,jingjiang@\}smu.edu.sg
\\
$^2$roy\_lee@sutd.edu.sg
}
\begin{document}

\maketitle

\begin{abstract}
Hateful meme classification is a challenging multimodal task that requires complex reasoning and contextual background knowledge. Ideally, we could leverage an explicit external knowledge base to supplement contextual and cultural information in hateful memes. However, there is no known explicit external knowledge base that could provide such hate speech contextual information. To address this gap, we propose \textsf{PromptHate}, a simple yet effective prompt-based model that prompts pre-trained language models (PLMs) for hateful meme classification. Specifically, we construct simple prompts and provide a few in-context examples to exploit the implicit knowledge in the pre-trained RoBERTa language model for hateful meme classification. We conduct extensive experiments on two publicly available hateful and offensive meme datasets. Our experimental results show that \textsf{PromptHate} is able to achieve a high AUC of 90.96, outperforming state-of-the-art baselines on the hateful meme classification task. We also perform fine-grained analyses and case studies on various prompt settings and demonstrate the effectiveness of the prompts on hateful meme classification. 
\end{abstract}

{\color{red} \textbf{Disclaimer}: \textit{This paper contains discriminatory content that may be disturbing to some readers.}} 

% Specifically, the examples used in the figures and tables contain actual examples of hateful memes and hate speech targeting particular groups. These examples are very offensive and distasteful. However, we have made the hard decision to display these actual hateful examples to provide context on the toxicity of malicious content that we are dealing with. Besides making technical contributions in this paper, we hope the distasteful examples used could also raise awareness of the vulnerable groups targeted in hate speeches in the real-world.

\section{Introduction}

\label{sec:intro}
\input{Introduction}

\section{Related Work}
\label{sec:related}
\input{Related}

\section{Preliminaries}
\label{sec:prelim}
\input{Preliminary}

\section{Methodology}
\label{sec:model}
\input{Model}

\section{Experiments}
\label{sec:exp}
\input{Experiments}
\section{Conclusion}
\label{sec:conclusion}
\input{Conclusion}

\section{Limitations}
\label{sec:limit}
\input{Limitation}

%\clearpage

\bibliographystyle{acl_natbib}
\bibliography{ref}

\clearpage
\section*{APPENDIX}
\appendix
\input{Appendix}

\end{document}

%% file: Introduction.tex
\begin{figure}[t] 
%	\centering
	\includegraphics[width=\linewidth]{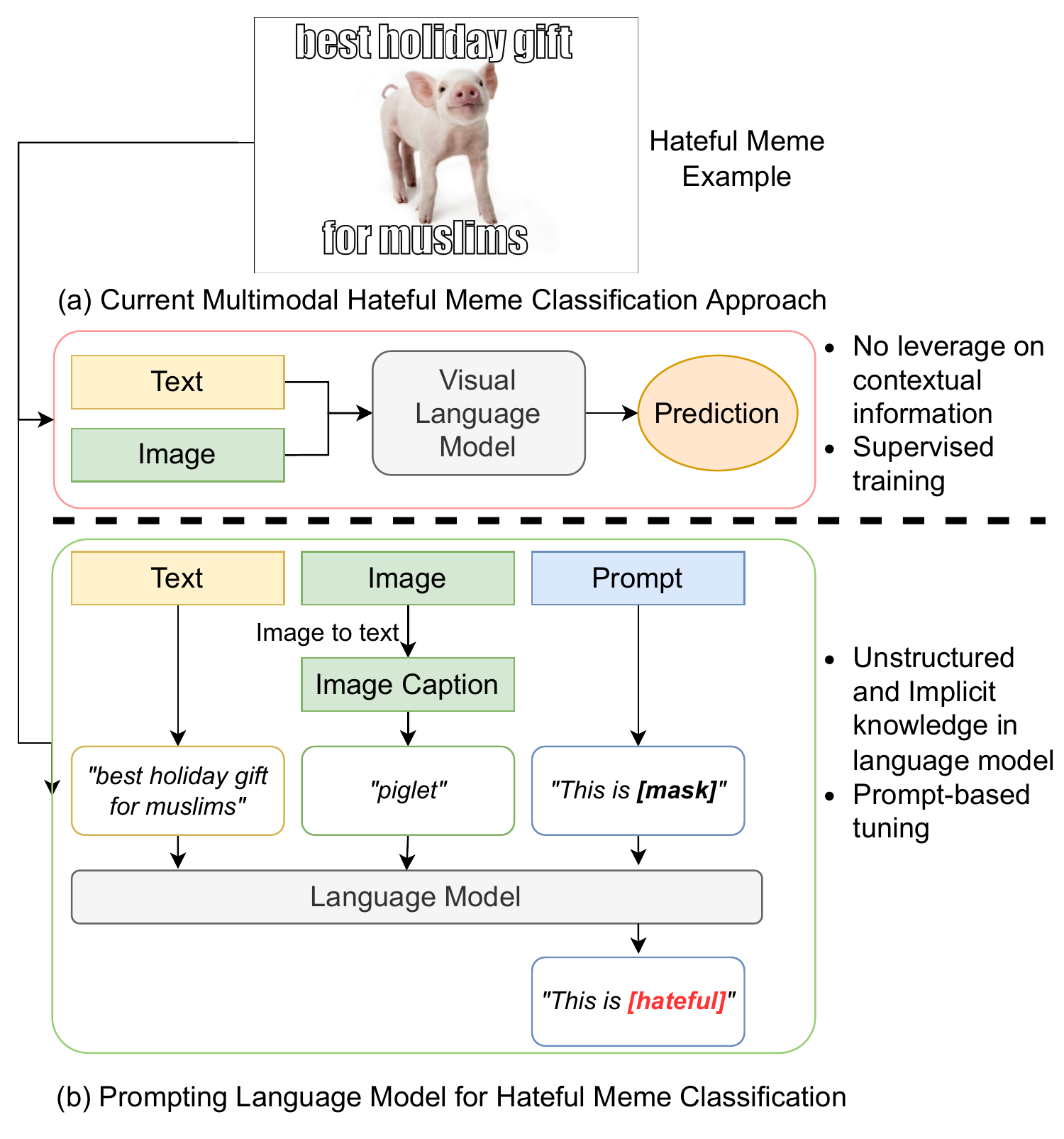} 
	\caption{Comparison between (a) fine-tuning visual language model approach and (b) prompt-based approach.} %\crcomment{This image should be changed cause the meme and the textual description in the image does not match}}
	\label{fig:intro-img}
\end{figure}

Internet memes have evolved into one of social media's most popular forms of communication. Memes are presented as images with accompanying text, which are usually intended to be funny or satirical in nature. However, malicious online users generate and share hateful memes under the guise of humor~\cite{DBLP:conf/nips/KielaFMGSRT20,DBLP:conf/acl/PramanickDMSANC21,DBLP:conf/wacv/GomezGGK20,DBLP:conf/acl-trac/SuryawanshiCAB20}. Many hateful memes had been created to attack and ridicule people based on specific characteristics such as race, ethnicity, and religion. The hateful memes could also threaten society more than text-based online hate speech due to their viral nature; users could re-post or share these hateful memes in multiple conversations and contexts. 

To combat the spread of hateful memes, social media platforms such as Facebook have recently released a large hateful meme dataset as part of a challenge to encourage researchers to develop automated solutions to perform hateful memes classification~\cite{DBLP:conf/nips/KielaFMGSRT20}. However, classifying hateful memes turns out to be a very challenging task as the solution would need to comprehend and reason across both the visual and textual modalities. The reasoning of the modalities will also require contextual background knowledge.

Recent research works have proposed multimodal hateful meme classification approaches. For instance, some studies have adopted pre-trained visual language models such as VilBERT~\cite{lu2019vilbert} and VisualBERT~\cite{li2019visualbert} and fine-tune these models with the hateful meme classification task~\cite{lippe2020multimodal,zhu2020enhance,zhou2020multimodal,DBLP:journals/corr/abs-2012-07788,DBLP:journals/corr/abs-2012-12975,DBLP:conf/websci/ZhuLC22} (as illustrated in Figure~\ref{fig:intro-img}(a)). Nevertheless, existing approaches still have limitations as understanding hateful memes may require additional contextual background knowledge. Consider the hateful meme example in Figure~\ref{fig:intro-img}. %The background knowledge that Muslim women wear hijab is required to infer that the meme is hateful. 
The background knowledge that the pig is considered unclean by Muslims and is a sin to consume, is required to infer that the meme is hateful.

%Some studies have utilized two-stream frameworks, which process images and texts independently and later fuse them using multimodal fusion techniques to train a hateful meme classsifier~\cite{DBLP:journals/corr/abs-2012-04977,DBLP:conf/nips/KielaFMGSRT20,DBLP:conf/acl-trac/SuryawanshiCAB20}. Other

%Ideally, we could leverage an explicit external knowledge base to supplement such contextual background knowledge for hateful meme classification. However, to the best of our knowledge, no such knowledge base exists. It is also costly and labour-intensive to construct such a hateful context knowledge base. Therefore, we explore leveraging pre-trained language models (PLMs) as unstructured and implicit knowledge bases~\cite{DBLP:journals/corr/abs-2109-05014,DBLP:journals/corr/abs-1806-02847,DBLP:conf/emnlp/PetroniRRLBWM19}.

Recent studies have attempted to prompt Pre-trained Language Models (PLM) and yield good performance for uni-modal NLP~\cite{DBLP:conf/eacl/SchickS21,DBLP:conf/nips/BrownMRSKDNSSAA20,DBLP:conf/naacl/SchickS21,DBLP:conf/acl/GaoFC20}.
%and computer vision tasks~\cite{DBLP:journals/corr/abs-2203-05557,DBLP:journals/corr/abs-2109-01134,DBLP:conf/icml/RadfordKHRGASAM21} \crcomment{I'll update this part.}. 
Nevertheless, few works have attempted to prompt PLMs for multimodal tasks~\cite{DBLP:journals/corr/abs-2109-11797,DBLP:journals/corr/abs-2204-00598,DBLP:journals/corr/abs-2112-08614}. \citet{DBLP:journals/corr/abs-2109-05014} has explored prompting GPT-3 model~\cite{DBLP:conf/nips/BrownMRSKDNSSAA20} for the visual question \& answering task. However, the approach has limitations as large models such as GPT-3 are expensive to tune. This study addresses the research gaps and proposes a novel framework to leverage the implicit and unstructured knowledge in PLMs~\cite{DBLP:journals/corr/abs-1806-02847,DBLP:conf/emnlp/PetroniRRLBWM19} to improve hateful meme classification. Figure~\ref{fig:intro-img} illustrates the comparison between the existing multimodal hateful meme classification approach and our proposed prompt-based approach. Specifically, in our prompt-based approach, we first convert images into textual descriptions that a PLM can understand and design specific prompts to adapt and leverage the implicit knowledge in the PLM. Subsequently, given a meme and a prompt, the prompt-tuned PLM generates a textual output, indicating the predicted label of the input meme. The underlying intuition for the prompt-based approach is that PLMs will tap into the implicit and unstructured knowledge in their large-scale pre-training data to generate the continuation of the prompt, i.e., from ``\textit{this is \_}'' to ``\textit{this is \textbf{hateful}.}''.

%{\color{red} Roy: Need to make sure that the recent prompting works are cited - especially those that are prompting visual-language models}

%PLMs are pre-trained with large-scale data, including contents from web~\cite{radford2019language}, data from books~\cite{DBLP:conf/iccv/ZhuKZSUTF15} and texts from Wikipedia. Therefore, these PLMs have captured vast knowledge and can be regarded as implicit and unstructured knowledge bases. 

%So far, this idea has mostly been considered in few-shot or zero-shot scenarios where there is little or no training data~\cite{DBLP:conf/acl/GaoFC20,DBLP:conf/eacl/SchickS21,DBLP:conf/naacl/SchickS21}. Existing studies have also demonstrated that performing prompt-based tuning on PLMs yielded superior classification performance over fine-tuning PLMs for the classification task, especially in few-shot settings~\cite{DBLP:conf/eacl/SchickS21,DBLP:conf/nips/BrownMRSKDNSSAA20,DBLP:conf/naacl/SchickS21,DBLP:conf/acl/GaoFC20}. In this paper, we propose \textsf{PromptHate}, which is a prompt-based model based on RoBERTa~\cite{DBLP:journals/corr/abs-1907-11692} for hateful meme classification. To the best of our knowledge, this is the first paper that proposes a prompt-based approach to perform hateful meme classification.

We summarize this paper's contribution as follows: (i) We propose a multimodal prompt-based framework called \textsf{PromptHate}, which prompts and leverages the implicit knowledge in the PLM to perform hateful meme classification. (ii) We conduct extensive experiments on two publicly available datasets. Our experiment results show that \textsf{PromptHate} outperforms state-of-the-art methods for the hateful meme classification task. (iii) We perform fine-grained analyses and case studies on various settings to examine the prompts' effectiveness in classifying hateful meme. To the best of our knowledge, this is the first paper that explore prompting PLM for hateful meme classification. \crtext{\footnote{Code: https://gitlab.com/bottle\_shop/safe/prompthate}}
%\crcomment{I've moved the related work into the last section of the paper: before the section of Limitation, as most papers do.} 

%% file: Related.tex
\subsection{Hateful Meme Detection}
Hateful meme classification is an emerging multimodal task made popular by the availability of several recent hateful memes datasets~\cite{DBLP:conf/nips/KielaFMGSRT20,DBLP:conf/acl-trac/SuryawanshiCAB20,DBLP:conf/wacv/GomezGGK20}. For instance, Facebook had organized the \textit{Hateful Memes Challenge}, which encouraged researchers to submit solutions to perform hateful memes classification~\cite{DBLP:conf/nips/KielaFMGSRT20}. The memes are specially constructed such that unimodal methods cannot yield good performance in this classification task. Therefore, existing studies have adopted multimodal approaches to perform hateful memes classification.

Existing studies have explored \textit{classic two-stream models} that combine the text and visual features learned from text and image encoders using attention-based mechanisms and other fusion methods to perform hateful meme classification~\cite{DBLP:journals/corr/abs-2012-04977,DBLP:conf/nips/KielaFMGSRT20,DBLP:conf/acl-trac/SuryawanshiCAB20}. Another popular line of approach is fine-tuning large scale pre-trained multimodal models for the task~\cite{lippe2020multimodal,zhu2020enhance,zhou2020multimodal,DBLP:journals/corr/abs-2012-07788,DBLP:journals/corr/abs-2012-12975,DBLP:conf/emnlp/PramanickSDAN021,hee2022explaining}. Recent studies have also attempted to use data augmentation~\cite{zhu2020enhance,zhou2020multimodal,DBLP:conf/websci/ZhuLC22} and ensemble methods~\cite{zhu2020enhance,DBLP:journals/corr/abs-2012-12975,DBLP:journals/corr/abs-2012-13235} to enhance the hateful memes classification performance. In a recent work, \citet{DBLP:conf/mm/LeeCFJC21} proposed the \textit{DisMultiHate} mode in attempted to disentangle hateful targets in memes. Nevertheless, existing studies may be inadequate in modeling the contextual background knowledge encoded in the hateful memes. This paper aims to fill this research gap by prompting the PLM to leverage its unstructured implicit knowledge for hateful meme classification.

\subsection{Language Model Prompting}
The increasing popularity of large-scale PLMs such as GPT~\cite{radford2019language,DBLP:conf/nips/BrownMRSKDNSSAA20}, BERT~\cite{devlin2018bert}, RoBERTa~\cite{DBLP:journals/corr/abs-1907-11692} has also popularized prompt-based learning. Existing prompt-based learning studies have explored using PLMs as implicit and unstructured knowledge bases~\cite{DBLP:journals/tacl/TalmorEGB20,DBLP:conf/emnlp/DavisonFR19,DBLP:conf/conll/SchwartzSKZCS17}. Recent studies have also prompted PLMs for various NLP tasks such as natural language inference and sentiment classification, and yield good performance in few-shot settings~\cite{DBLP:conf/acl/GaoFC20,DBLP:conf/eacl/SchickS21,DBLP:conf/naacl/SchickS21}. There are also recent works that prompt visual-language models for computer vision tasks ~\cite{DBLP:journals/corr/abs-2203-05557,DBLP:journals/corr/abs-2109-01134,DBLP:conf/icml/RadfordKHRGASAM21}.

Nevertheless, most of the existing prompt-based learning studies are limited to unimodal tasks, and there are fewer works on prompting PLM for multimodal tasks~\cite{DBLP:journals/corr/abs-2109-11797,DBLP:journals/corr/abs-2204-00598,DBLP:journals/corr/abs-2112-08614}. \citet{DBLP:journals/corr/abs-2109-05014} has explored prompting GPT-3 model~\cite{DBLP:conf/nips/BrownMRSKDNSSAA20} for the visual question \& answering task. However, there are limitations: large models such as GPT-3 are expensive to tune. Furthermore, the constraint on the input length limits the number of training instances. In this paper, we adopt a different approach and propose a novel framework, \textsf{PromptHate}, which prompts RoBERTa~\cite{DBLP:journals/corr/abs-1907-11692} for the multimodal hateful meme classification. \textsf{PromptHate} is a much smaller model compared to GPT-3, and it can be fine-tuned with training instances.

%% file: Preliminary.tex
\subsection{Problem Definition}
We define the problem of multimodal hateful memes classification as follows: Given a meme with image $\mathcal{I}$ and text $\mathcal{O}$, a classification model will predict the label of the multimodal meme (\textit{hateful} or \textit{non-hateful}). Traditionally, this binary classification task requires models to predict a probability vector $\mathbf{y} \in \mathbb{R}^2$ over the two classes. Specifically, $y_0$ denotes the predicted probability that the meme is non-hateful while $y_1$ is for the probability that the meme is hateful. If $y_1 > y_0$, the meme is predicted as hateful, otherwise, non-hateful. In framework, we transform the hateful meme classification task into a Masked Language Modelling (MLM) problem. Specifically, a PLM is prompted to replace the \texttt{[MASK]} token that represents the label of the meme (e.g., hateful or non-hateful). We discuss the prompting details in Section~\ref{sec:prompts-gen}.

\subsection{Image Captioning}
\label{sec:imagecaption}
To prompt PLMs for multimodal hateful meme classification, we first need to covert the meme's image into an acceptable textual input for PLMs. A common approach to extract the image's semantics and represent it with textual description is via image captioning~\cite{DBLP:journals/corr/abs-2109-05014,DBLP:journals/corr/abs-2112-08614}. We first extract the text in the memes using open-source Python packages EasyOCR\footnote{https://github.com/JaidedAI/EasyOCR}, followed by in-painting with  MMEditing\footnote{https://github.com/open-mmlab/mmediting} to remove the text. We then apply a pre-trained image captioning model, ClipCap~\cite{DBLP:journals/corr/abs-2111-09734}. ClipCap is able to generate good quality captions for low-resolution web images. The generated captions tend to describe the dominant objects or events in the meme's image and we use these captions as inputs into the \textsf{PromptHate} model. 

Besides captioning the image, we also leveraged Google Vision Web Entity Detection API\footnote{https://cloud.google.com/vision/docs/detecting-web} and pre-trained FairFace classifier~\cite{karkkainen2019fairface} to extract the entities in the memes and the demographic information if the meme contains a person. The extracted entities and demographic information are used as supplementary information that will be combined with the image captions as input to the PLMs. Note that although the extracted supplementary information may capture key information about the meme, the contextual background knowledge is still absent in the image caption and supplementary information. For instance, with the utilization of entity information, we may identify a pig in the meme and extract the term ``\textit{Muslim}'' from the meme text. However, the contextual knowledge that Muslims do not eat pork is absent in the supplementary information. %Therefore, we design a framework to prompt and leverage implicit knowledge in PLMs.

%% file: Model.tex
\begin{figure*}[t] 
	\centering
	\includegraphics[scale = 0.6]{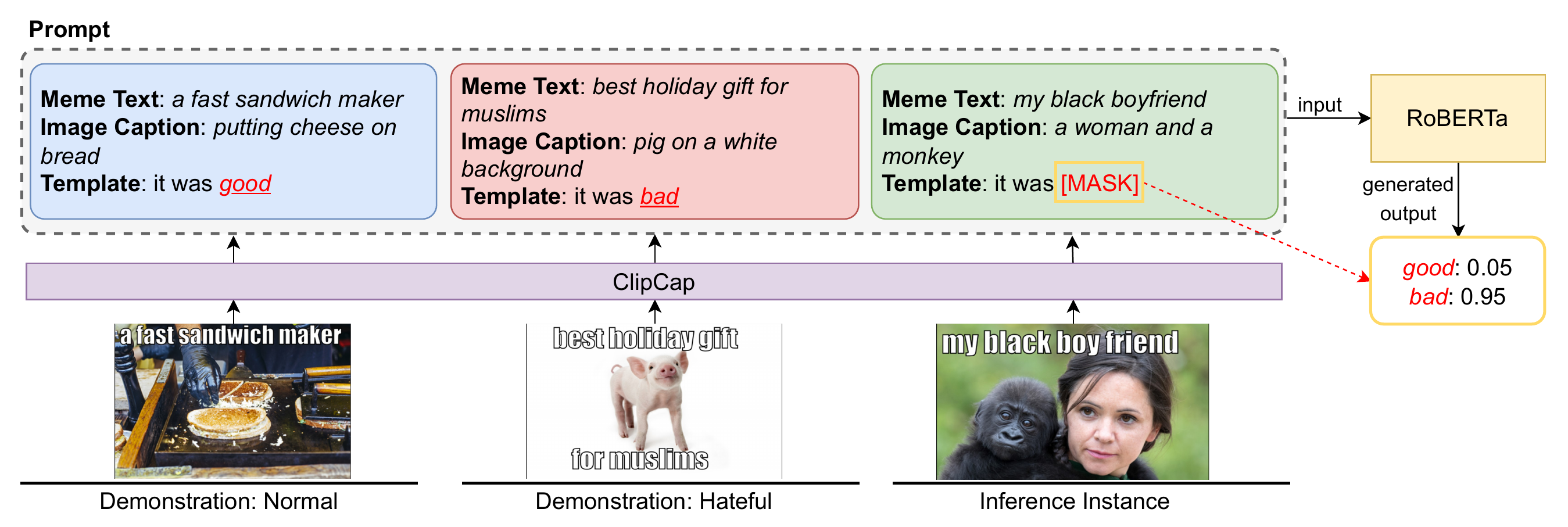}
	\caption{Overview of \textsf{PromptHate} Framework.}
	\label{fig:architecture}
\end{figure*}

Figure~\ref{fig:architecture} illustrates the architectural framework of our proposed \textsf{PromptHate} model. A key step in the \textsf{PromptHate} is the construction of a prompt, which consists of a positive demonstration (i.e., normal meme), a negative demonstration (i.e., hateful meme), and an inference instance (i.e., meme to be predicted). We first convert the three memes into meme texts and image descriptions using the data pre-processing steps described in Section~\ref{sec:imagecaption}. Subsequently, we construct templates, which are natural sentences that include label words for the individual memes. For instance, a normal demonstration meme will have a template ``\textit{the meme is \textbf{good}}'', while the hateful demonstration meme uses the template ``\textit{the meme is \textbf{bad}}''. The label word in template for the inference instance is replaced with a \textbf{[MASK]},  which the PLM (i.e., RoBERTa) is tasked to complete the sentence with ``\textit{\textbf{good}}'' or ``\textit{\textbf{bad}}''. Subsequent sections provide the technical details on prompting for the multimodal hateful meme classification task.
% \crcomment{seems we give the definition of label words in section 3.2, but we've used it here and section 3.1.}
%\crcomment{The second reviewer said that it would be a bit confusing cause previously we use ``hateful'' while here we use ``bad''. Shall we give some explainations or leave it there.}

\subsection{Prompting Hateful Meme}
\label{sec:prompts-gen}

To guide the PLM in inferring the label word, we also provide \textit{positive} and \textit{negative} demonstrations to the PLM. The positive demonstration $\mathcal{S}^{\text{pos}}$ is generated as: $\mathcal{S}_1^{\text{pos}}\texttt{[SEP]}\mathcal{S}_2^{\text{pos}}\texttt{[SEP]}\mathcal{T}(\mathcal{W}_{\text{pos}})$, where $\mathcal{S}_1^{\text{pos}}$ and $\mathcal{S}_2^{\text{pos}}$ are meme texts and image descriptions respectively, \texttt{[SEP]} is the separation token in the language model $\mathcal{L}$, and $\mathcal{T}(\mathcal{W}_{\text{pos}})$ generates the positive label word $\mathcal{W}_{\text{pos}}$ into a sentence (e.g., ``\textit{this is \textbf{good}}'').
%\crcomment{should we clarify what $\mathcal{S}_1^{\text{pos}}$ and $\mathcal{S}_2^{\text{pos}}$ mean?} 
Similar approach is used for the generation of negative demonstration $\mathcal{S}^{\text{neg}}$ and inference instance $\mathcal{S}^{\text{infer}}$ by replacing $\mathcal{W}_{\text{pos}}$ with $\mathcal{W}_{\text{neg}}$ and \texttt{[MASK]}, respectively. Inspired by \citet{DBLP:conf/acl/GaoFC20}, we concatenate the demonstrations with the inference instance:

\begin{equation}
    \mathcal{S} = \texttt{[START]}\mathcal{S}^{\text{infer}} \texttt{[SEP]} \mathcal{S}^{\text{pos}} \texttt{[SEP]} \mathcal{S}^{\text{neg}} \texttt{[END]}
\end{equation}
where, $\mathcal{S}$ serves as the prompt fed into $\mathcal{L}$, and \texttt{[START]} and \texttt{[END]} are start and end tokens in $\mathcal{L}$.

\subsection{Templates and Label Words}
Recent studies have explored designing better prompts by developing automatic template generation and label word selection methods~\cite{DBLP:conf/acl/GaoFC20}. As \textsf{PromptHate} is the first study that adopted prompting for hateful meme classification, we adopt a simpler approach of prompting with manually defined label words and templates.

Labels are required to be mapped into individual words for prompt-based models. As shown in Figure~\ref{fig:architecture}, \textit{good} is used as the label word for the positive class (non-hateful), while \textit{bad} for the negative class (hateful). We have also analysed sets of other label words. The comparison of using different label words is discussed in Section~\ref{sec:experiment-results}.

The template in prompts can be viewed as the function $\mathcal{T}$, which maps the label word into a sentence. In \textsf{PromptHate}, we manually define the function $\mathcal{T}(\texttt{[WORD]}) \to \textit{It was \texttt{[WORD]}.}$. Specifically, if $\mathcal{T}$ receives $\mathcal{W}_{\text{pos}}$ as input, the output sentence should be ``\textit{\textbf{It was $\mathcal{W}_{\text{pos}}$.}}''. Conversely, if $\mathcal{T}$ receives $\mathcal{W}_{\text{neg}}$ as input, the output sentence should be ``\textit{\textbf{It was $\mathcal{W}_{\text{neg}}$.}}''.

\subsection{Model Training and Prediction}
\label{sec:model-details}

For training, we feed the prompt $\mathcal{S}$ into $\mathcal{L}$ and obtain the probability of the masked word, $\mathbf{y} \in \mathbb{R}^2$ over label words:
\begin{align}
    y_0 &= \text{P} (\texttt{[MASK]}=\mathcal{W}_{\text{pos}}| \mathcal{S}), \\
    y_1 &= \text{P} (\texttt{[MASK]}=\mathcal{W}_{\text{neg}}| \mathcal{S}). 
\end{align}

The training loss is based on cross-entropy loss with the ground-truth label $\mathbf{\hat{y}}$:
\begin{equation}
    \text{Loss} = y_0log(\hat{y_0})+y_1log(\hat{y_1}),
\end{equation}
and the loss will be used for updating parameters $\bm{\theta}$ in $\mathcal{L}$. Differing from standard fine-tuning PLMs by adding a task-specific classification head, prompt-based tuning does not have additional parameters beyond those in the PLMs, and the MLM task does not deviate from PLM's pre-training objectives. 

For model prediction, we obtain the probability of the masked word over label words in the same manner. If $y_1>y_0$, the meme will be predicted as hateful, otherwise, non-hateful.

\subsection{Multi-Query Ensemble}
\label{sec:ensemble}
%\crcomment{According to the reviewer's comment (R1), I've changed the section from 4.3.1 to 4.4 cause we do not have 4.3.2}
Demonstrations in the prompt provide additional cues for the inference instance. Existing works carefully select demonstrations which are similar to the inference instance~\cite{DBLP:journals/corr/abs-2109-05014,DBLP:conf/acl/GaoFC20}. Nevertheless, memes that are similar in visual or textual modality may be targeting different protected characteristics (e.g., race, religion, gender, etc.), and understanding the target in the hateful meme is critical to the classification task~\cite{DBLP:conf/mm/LeeCFJC21}. To address this concern, we adopt a multi-query ensemble strategy to predict the inference instance using multiple pairs of demonstrations. Specifically, when we adopt a $M$-query ensemble, an inference instance will be predicted using $M$ pairs of demonstrations.

The multi-query ensemble will result in a set of prediction scores for the inference instance: $\{\mathbf{y}_m\}_{m=1}^M$, where $\mathbf{y}_m \in \mathbb{R}^2$ is the predicted scores with the $m$-th pair of demonstration. The final prediction will be the average over all predicted scores:
\begin{equation}
    \mathbf{y}_{\text{final}} = \frac{1}{M} \sum_{m=1}^M \mathbf{y}_m.
\end{equation}

%We adopt the AdamW algorithm~\cite{loshchilov2018fixing} for optimizing the parameters of the model. %, and the learning rate is set to be $10^{-5}$. The size of the mini-batch is set to be $16$. \crcomment{I think we can move this to the Appendix: experiment setting.}

%% file: Experiments.tex
In this section, we first provide a brief introduction to the datasets and evaluation setting. Next, we present a set of experiments conducted to evaluate \textsf{PromptHate}'s hateful meme classification performance. We also conduct studies to understand the effects of various prompt settings, and discuss the limitations of our model via error case studies. 

\subsection{Evaluation Settings}
%\subsection{Datasets}
\textbf{Datasets.} We used two publicly available datasets in our experiments: the \textit{Facebook Hateful Meme} dataset (FHM)~\cite{DBLP:conf/nips/KielaFMGSRT20} and the \textit{Harmful Meme} dataset (HarM)~\cite{DBLP:conf/acl/PramanickDMSANC21}. Table~\ref{tab:dataset} outlines the statistical distributions of the two datasets. The FHM dataset was constructed and released by Facebook as part of a challenge to crowd-source multimodal hateful meme classification solutions.We do not have labels of the memes in the test split. Therefore, we utilize the \textit{dev-seen} split as the \textit{test}. Due to the limited availability of public multi-modal hateful meme datasets, we choose HarM dataset containing misinformation memes as the other evaluation dataset.
The HarM dataset was constructed with real COVID-19-related memes collected from Twitter. The memes are labeled with three classes: \textit{very harmful}, \textit{partially harmful}, and \textit{harmless}. We combine the \textit{very harmful} and \textit{partially harmful memes} into hateful memes and regard harmless memes as non-hateful memes. The good performance on the HarM dataset also implies the generalization of the \textsf{PromptHate} to other anti-social memes besides hateful ones. %\crcomment{According the second reviewer's comment, it'd be a bit weird to combine harmful ones into the hateful class. Shall we address the issue here?}

%\textbf{Facebook Hateful Meme (FHM)}: The dataset was constructed and released by Facebook as part of a challenge to crowd-source multimodal hateful meme classification solutions. The dataset contains $10K$ memes with binary labels (i.e., hateful or non-hateful). We do not have labels of the memes in the test split. Therefore, we utilize the \textit{dev-seen} split as the \textit{test}.

%\textbf{Harmful Meme (HarM)}: The dataset is constructed with real COVID-19 related memes collected from Twitter. The memes are labeled with three classes: \textit{very harmful}, \textit{partially harmful}, and \textit{harmless}. We combine the \textit{very harmful} and \textit{partially harmful memes} into hateful memes and regard harmless memes as non-hateful memes.

\begin{table}[t]
  \begin{small}
  \begin{tabular}{c|cc|cc}
    \hline
    \textbf{Datasets}& \multicolumn{2}{c|}{\textbf{Train}} & \multicolumn{2}{c}{\textbf{Test}}\\
    & \#Hate. & \#Non-hate. & \#Hate. & \#Non-hate.\\
    \hline\hline
    FHM & 3,050 & 5,450 &  250 & 250 \\
    HarM & 1,064 & 1,949 & 124 & 230\\
    \hline

\end{tabular}
\end{small}
\caption{Statistical summary of FHM and HarM.}
  \label{tab:dataset}
\end{table}

%\subsection{Evaluation Settings}
%\subsubsection{Evaluation Metrics}
\textbf{Evaluation Metrics.} We adopt the evaluation metrics commonly used in existing hateful meme classification studies \cite{DBLP:conf/nips/KielaFMGSRT20,zhu2020enhance,zhou2020multimodal,DBLP:journals/corr/abs-2012-07788,DBLP:journals/corr/abs-2012-12975}: Area Under the Receiver Operating Characteristic curve (AUROC) and Accuracy (Acc).  In order to report more reliable results, we measure the average performance of models under \textbf{ten} random seeds. All models use the same set of random seeds.

%\subsubsection{Baselines}
\textbf{Baselines.} We benchmark \textsf{PromptHate} against the state-of-the-art hateful meme classification models. Specifically, we compare with two types of baselines models: (a) uni-modal models that only use information from one modality (i.e., the meme text or the meme image); (b) multimodal models. 

For uni-modal baselines, we consider a text-only model by fine-tuning pre-trained BERT on the meme text for classification (\textbf{Text BERT}). We also apply an image-only model, which processes the meme image using Faster R-CNN~\cite{ren2016faster} with ResNet-152~\cite{DBLP:conf/cvpr/HeZRS16} before feeding the image representation into a classifer for hateful meme classification (\textbf{Image-Region}).

For multimodal baselines, we compare with the multimodal methods benchmarked in the original FHM dataset paper~\cite{DBLP:conf/nips/KielaFMGSRT20}, namely: \textbf{Late Fusion}, \textbf{Concat BERT}, \textbf{MMBT-Region}~\cite{DBLP:conf/nips/KielaBFT19}, \textbf{ViLBERT CC}~\cite{lu2019vilbert}, \textbf{Visual BERT COCO}~\cite{li2019visualbert}. We also compare to the state-of-the-art hateful meme classification methods~\footnote{ Note that we use the code published by the author and re-run the model for ten rounds with different random seeds.}: \textbf{CLIP BERT}, \textbf{MOMENTA}~\cite{DBLP:conf/emnlp/PramanickSDAN021} and \textbf{DisMultiHate}~\cite{DBLP:conf/mm/LeeCFJC21}. CLIP BERT and MOMENTA are models leveraging image features generated by the CLIP model~\cite{DBLP:conf/icml/RadfordKHRGASAM21}. CLIP is pre-trained with web data, thus it is able to generalize well to hateful meme detection where images and texts are noisy. CLIP BERT uses CLIP as the visual encoder and BERT as the text encoder and feed the concatenation of features to a classifier for prediction. MOMENTA considers the global and local information in two modalities by modeling the deep multi-modal interactions. DisMultiHate disentangles target information from the meme to improve the hateful content classification.

As \textsf{PromptHate} prompts RoBERTa~\cite{DBLP:journals/corr/abs-1907-11692} for hateful meme classification, we also benchmark \textsf{PromptHate} against fine-tuning RoBERTa (\textbf{FT-RoBERTa}). Specifically, we concatenate the meme text and image descriptions as input to fine-tune RoBERTa, and the output representation is fed into a MLP layer for classification.

\subsection{Experiment Results}
\label{sec:experiment-results}

\begin{table}[t]
  \small
  \centering
  \begin{tabular}{c|cc}
    \hline
    \textbf{Model} & \textbf{AUC.} & \textbf{Acc.} \\
    \hline\hline
    Text BERT & 66.10$_{\pm0.55}$& 57.12$_{\pm0.49}$  \\
    Image-Region & 56.69$_{\pm1.05}$ &52.34$_{\pm1.39}$  \\
    \hline
    Late Fusion & 66.34$_{\pm1.54}$ & 59.14$_{\pm0.91}$\\
    Concat BERT & 66.53$_{\pm0.75}$&60.80$_{\pm0.98}$  \\
    MMBT-Region & 72.86$_{\pm0.64}$&65.06$_{\pm1.76}$  \\
    Visual BERT COCO & 68.71$_{\pm1.02}$& 61.48$_{\pm1.19}$ \\
    ViLBERT CC & 73.05$_{\pm0.62}$& 64.70$_{\pm1.12}$\\
    CLIP BERT & 66.97$_{\pm0.34}$& 58.28$_{\pm0.63}$ \\
    MOMENTA & 69.17$_{\pm4.71}$ & 61.34$_{\pm4.89}$ \\
    DisMultiHate &79.89$_{\pm1.71}$ & 71.26$_{\pm1.66}$  \\
    \hline
    FT-RoBERTa & 76.32$_{\pm6.45}$ &67.72$_{\pm6.20}$ \\
    \textsf{PromptHate} &\textbf{81.45$_{\pm0.74}$} &\textbf{72.98$_{\pm1.09}$ } \\
    \hline
\end{tabular}
\caption{Experimental results of models on FHM.}
\label{tab:exp-results-fhm}
\end{table}

\begin{table}[t]
  \small
  \centering
  \begin{tabular}{c|cc}
    \hline
    \textbf{Model} & \textbf{AUC.} & \textbf{Acc.} \\
    \hline\hline
    Text BERT & 81.39$_{\pm0.91}$& 75.68$_{\pm1.59}$\\
    Image-Region &76.46$_{\pm0.47}$ &73.05$_{\pm1.80}$  \\
    \hline
    Late Fusion & 83.17$_{\pm1.25}$& 77.57$_{\pm0.96}$\\
    Concat BERT & 83.21$_{\pm1.37}$& 77.82$_{\pm1.09}$  \\
    MMBT-Region & 85.48$_{\pm0.75}$& 79.83$_{\pm2.00}$  \\
    Visual BERT COCO & 80.46$_{\pm1.04}$&75.31$_{\pm1.44}$ \\
    ViLBERT CC & 84.11$_{\pm0.88}$ &78.70$_{\pm1.17}$ \\
    CLIP BERT  & 82.63$_{\pm1.20}$ & 76.66$_{\pm1.02}$ \\
    MOMENTA & 86.32$_{\pm3.83}$& 80.48$_{\pm1.95}$\\
    DisMultiHate & 86.39$_{\pm1.17}$ &81.24$_{\pm1.04}$  \\
    \hline
    FT-RoBERTa &89.26$_{\pm1.04}$ & 82.32$_{\pm1.60}$ \\
    \textsf{PromptHate} & \textbf{90.96$_{\pm0.62}$} & \textbf{84.47$_{\pm1.75}$}  \\
    \hline
\end{tabular}
\caption{Experimental results of models on HarM.}
\label{tab:exp-results-harm}
\end{table}
%{\color{red} (WH: For FHM, FT-RoBERTa is not the best baseline. DisMultiHate is. We need to significance test against DisMultiHate.)}
Table~\ref{tab:exp-results-fhm} and \ref{tab:exp-results-harm} show the experimental results on FHM and HarM datasets, respectively. The standard deviations ($\pm$) of the ten runs are also reported, and the best results are \textbf{bold}. \textsf{PromptHate} outperforms the state-of-the-art baselines in both datasets. We have also computed the statistical differences between \textsf{PromptHate} and the best-performing baseline (i.e., DisMultiHate on FHM and FT-RoBERTa on HarM), and \textsf{PromptHate}'s improvement over the baseline is found to be statistically significant (\textit{p-value} < 0.05). Consistent with the existing studies, the multimodal approaches outperformed the unimodal baselines. More interestingly, we noted \textsf{PromptHate}'s improvements over the multimodal baselines that fine-tuned PLMs and FT-RoBERTa, demonstrating the strength of the prompting approach for the hateful meme classification task. Specifically, the performance comparison of FT-RoBERTa and \textsf{PromptHate} suggests that the prompting approach can better leverage the implicit knowledge embedded in the PLM  by adopting a masked language modeling training objective for the hateful meme classification.

We also observe differences in \textsf{PromptHate}'s performance on the FHM and HarM datasets; the model yields better performance on HarM. Similar observations are made for the other models. We postulate that the performance differences are likely due to the difficulty of the dataset. FHM contains hateful memes on multiple topics, while HarM mainly contains COVID-19-related hateful memes. Therefore, the models would have to be able to generalize better to perform well on the FHM dataset. We also highlight the high standard deviation in FT-RoBERTa's performance on FHM, suggesting FT-RoBERTa's instability and difficulty in generalizing well on the dataset.

As RoBERTa-large is regarded as a general LM for prompting~\cite{DBLP:conf/acl/GaoFC20,DBLP:conf/naacl/SchickS21,DBLP:conf/eacl/SchickS21}, \textsf{PromptHate} with RoBERTa-large is three times in the scale compared with BERT-base related baselines. To further valid the effectiveness of prompting approach in hateful meme detection, we conduct the following experiments: 1) we replace the RoBERTa-large with RoBERTa-base in \textsf{PromptHate} (\textsf{PromptHate-RB}); 2) we replace the BERT-base in the baseline models with either RoBERTa-large (\textbf{-RL}) or BERT-large (\textbf{-BL}). Specifically, we choose the most powerful baseline, DisMultiHate, for analysis. Experimental results on FHM and HarM are summarized in Table~\ref{tab:exp-comp-fhm} and Table~\ref{tab:exp-comp-HarM} respectively, where each block includes models of similar sizes.

\begin{table}[t]
  \small
  \centering
  \begin{tabular}{c|cc}
    \hline
    \textbf{Model} & \textbf{AUC.} & \textbf{Acc.} \\
    \hline\hline
     DisMultiHate &\textbf{79.89}$_{\pm1.71}$ & \textbf{71.26}$_{\pm1.66}$  \\
    \textsf{PromptHate-RB} &79.17$_{\pm0.67}$ &70.56$_{\pm0.73}$  \\
    \hline
    DisMultiHate-BL&79.97$_{\pm1.19}$ & 71.62$_{\pm1.15}$  \\
    DisMultiHate-RL&78.56$_{\pm0.94}$ & 71.10$_{\pm1.58}$  \\
    \textsf{PromptHate} &\textbf{81.45$_{\pm0.74}$} &\textbf{72.98$_{\pm1.09}$ } \\
    \hline
\end{tabular}
\caption{Experimental results of models on FHM.}
\label{tab:exp-comp-fhm}
\end{table}

\begin{table}[t]
  \small
  \centering
  \begin{tabular}{c|cc}
    \hline
    \textbf{Model} & \textbf{AUC.} & \textbf{Acc.} \\
    \hline\hline
    DisMultiHate & 86.39$_{\pm1.17}$ &81.24$_{\pm1.04}$  \\
    \textsf{PromptHate-RB} &\textbf{89.20}$_{\pm0.72}$ &\textbf{83.70}$_{\pm1.99}$ \\
    \hline
    DisMultiHate-BL&85.38$_{\pm1.13}$ & 80.71$_{\pm1.45}$  \\
    DisMultiHate-RL&88.39$_{\pm0.74}$ & 82.18$_{\pm1.13}$  \\
    \textsf{PromptHate} & \textbf{90.96$_{\pm0.62}$} & \textbf{84.47$_{\pm1.75}$}  \\
    \hline
\end{tabular}
\caption{Experimental results of models on HarM.}
\label{tab:exp-comp-HarM}
\end{table}

Unsurprisingly, replacing the RoBERTa-large with RoBERTa-base worsens PromptHate performance. However, we do observe that \textsf{PromptHate-RB} still outperforms DisMultiHate on the HarM dataset. On the FHM dataset, \textsf{PromptHate-RB} has performed slightly worse than DisMultiHate but depicted higher stability regarding the standard deviation. 
Interestingly, replacing the text encoder of DisMultiHate with a larger pre-trained LM does not outperform \textsf{PromptHate} on both datasets. From the experimental results, we observe that model size plays a critical role in \textsf{PromptHate} performance. Nevertheless, the experimental results have also demonstrated the effectiveness of our proposed prompting approach over state-of-the-art baselines. 

\subsection{Ablation Study}
Table~\ref{tab:ablations} shows the ablation analysis of \textsf{PromptHate}. We notice removing the MLM training objective decreases \textsf{PromptHate}'s performance significantly. The MLM training objective is designed to align with the PLMs' training objectives. This plays a significant role in enabling \textsf{PromptHate} to better utilize the embedding implicit knowledge in the PLMs for hateful meme classification. Interestingly, we observe that \textsf{PromptHate} can perform well even without the demonstrations. Nevertheless, the effects of demonstrations in prompt-based model remains an open research topic, which requires further studies ~\cite{DBLP:journals/corr/abs-2202-12837}.
%This seems to be counter-intuitive and differ from existing studies and more research will need to be conducted to evaluate the role of demonstration in prompt-based hateful meme classification.       

%In this section, we conduct analysis about which part in \textsf{PromptHate} helps in hateful meme detection. Compared with standard fine-tuning, \textsf{PromptHate} has two major differences: 1) adapt the objective of masked language modelling (MLM) which is in consistency of PLMs' pre-training objectives; 2) concatenate demonstrations with the inference instance to provide contextual information. To analyze the importance of each component, we ablate them and illustrate performances in Table~\ref{tab:ablations}.
%According to the results, we can observe that both MLM and demonstrations helps. Specifically, the MLM objective helps more in hateful meme detection. It indicates that being consistent with the pre-training objective may facilitate in utilizing embedded knowledge in PLMs so that enhancing the performance of hateful meme detection. On the other hand, demonstrations provides hints about how the \texttt{[MASK]} token should be retrieved in the inference instance. On HarM dataset, the improvements of using demonstrations are minor. However, on more challenging FHM, the improvements are more obvious. 

\begin{table}[t]
\small
\centering
 \begin{tabular}{c|cc|cc}
    \hline
    %\multirow{2}{*}{\textbf{Setting}} &
    \textbf{Setting}& \multicolumn{2}{c|}{\textbf{FHM}} & \multicolumn{2}{c}{\textbf{HarM}}\\

     & AUC. & Acc. & AUC. & Acc.\\
         \hline\hline
    %w/o MLM and Demo.& 76.32 &67.72&89.26 & 82.32\\
    PromptHate &81.45 &72.98 &90.96 &84.47  \\
    w/o MLM &76.32 & 67.72 & 89.26 & 82.32\\
        w/o Demo. & 80.37 & 71.76 & 90.38 & 84.35\\
    %\hline
    %\hline\hline
    %\multirow{6}{*}{Few-Shot} & Normal & Hate &69.21 &63.88&86.45&72.69 \\
    %&Hate & Normal &62.17 &57.56&84.79&72.04 \\
    %&Benign & Offensive &68.91 &63.68&83.29&62.91 \\
    %&Offensive & Benign &64.80 &59.44&79.20&61.75 \\ 
    %&Good & Bad & 69.30&63.76&84.78&74.63 \\
    %&Bad & Good &67.40 &61.64&83.68&68.03 \\ 
\hline
\end{tabular}
\caption{Ablation study of \textsf{PromptHate}.}
  \label{tab:ablations}
\end{table}
 
\subsection{Prompt Engineering} 
Designing good prompts is essential to prompt-based models. In this section, we discuss how varying the prompts affect \textsf{PromptHate}'s performance in hateful meme classification. 

\subsubsection{Engineering Label Words}
Label words are individual words representing the labels used in prompt generation. We investigate the effects of replacing the prompts in \textsf{PromptHate} with different sets of label words. Specifically, we replace the label words in the prompt's positive and negative demonstrations in our experiments. Table~\ref{tab:verbalizer} presents the results. For example, in the first row in Table~\ref{tab:verbalizer}, we use ``\textit{It was \textbf{normal}}'' for positive demonstrations (i.e., non-hateful memes), and ``\textit{It was \textbf{hate}}'' for negative demonstration (i.e., hateful memes). Intuitively we aim to examine how the label word's semantics affect hateful meme classification. For a more extensive investigation, we conduct the experiments on full training and few-shot setting, i.e., using only 10\% of training instances.

%Conversely, in the second row, we swap the label words and use `\textit{It was \textbf{hate}}'' for positive demonstrations (i.e., non-hateful memes), and ``\textit{It was \textbf{normal}}'' for negative demonstration (i.e., hateful memes). The intuition for the experimental design is that we aim to examine how the label word's semantic affects hateful meme classification memes. For instance, will assigning a positive word (e.g., normal) to be the label word for negative demonstrations (i.e., hateful memes) hamper the \textsf{PromptHate}'s performance? 
%We have also conducted the experiments on full training instances and few-shot setting, i.e., using only 10\% of training instances.
%\crcomment{The difference of using different label words may not be obvious. Will the reviewers argue that how will this happen. Or we just give some plausible reasons in advance. I tried to give some explanations below.} 

Table~\ref{tab:verbalizer} shows that different prompts can lead to substantial differences in performances. Specifically, label words aligned to the semantic classes are able to achieve better performance compared to the reverse mapping (i.e., the last row of each setting). Interestingly, the differences between the semantic class-aligned prompts and the reverse mapping are more significant in the few-shot setting. A possible reason could be in the few-shot setting, the \textsf{PromptHate} relies more on the label words' semantics to extract implicit knowledge for hateful meme classification. Thus, the label words with the aligned semantic class will provide better context in the prompt to improve hateful meme classification when there are insufficient observations in training instances. Conversely, when \textsf{PromptHate} is trained with enough instances, the representations of the label words are updated to be closer to the hateful meme classification task.

\begin{table}[t]
\small
\centering
  \begin{tabular}{ccc|c|c}
    \hline
    \multirow{2}{*}{\textbf{Setting}} &
    \multicolumn{2}{c|}{\textbf{Label Words}}& 
    \multicolumn{2}{c}{\textbf{FHM}} \\
    %\multicolumn{2}{c|}{\textbf{FHM}} 
    %\multicolumn{2}{c}{\textbf{HarM}}\\

   & Pos. & Neg. & AUC & Acc\\%AUC. & Acc. & AUC. & Acc.\\
    \hline\hline
        \multirow{4}{*}{full} 
    %& Normal & Hate &81.21 &71.74   &90.45 &84.66  \\
    & Normal & Hate &81.21 &71.74   \\
    & Benign & Offensive & 81.58 &72.70 \\
    %&Benign & Offensive & 81.58& 72.70 &89.66 &82.83  \\
%    Offensive & Benign & 81.00 &72.46  &89.70 &83.81  \\ 
    %&Good & Bad &81.45 &72.98 &90.96 &84.47 \\
    &Good & Bad &81.45 &72.98 \\
    %&Bad & Good & 81.12& 73.12 &89.67 &83.22  \\ 
    %&Hate & Normal & 80.51 &72.22  &89.81 &83.22  \\
    &Hate & Normal & 80.51  &72.22  \\
    %\hline
    \hline
    \multirow{4}{*}{Few-Shot} 
    %& Normal & Hate &69.21 &63.88&86.45&72.69 \\
    & Normal & Hate &69.21&63.88 \\
    
    %&Benign & Offensive &68.91 &63.68&83.29&62.91 \\
    &Benign & Offensive &68.91&63.68 \\
    %&Offensive & Benign &64.80 &59.44&79.20&61.75 \\ 
    %&Good & Bad & 69.30&63.76&84.78&74.63 \\
    &Good & Bad & 69.30&63.76 \\
    %&Bad & Good &67.40 &61.64&83.68&68.03 \\ 
    %&Hate & Normal &62.17 &57.56&84.79&72.04 \\
    &Hate & Normal &62.17 &57.56  \\
\hline
\end{tabular}
\caption{\textsf{PromptHate} with various label words.}
  \label{tab:verbalizer}
\end{table}

\subsubsection{Prompt with Hateful Target Information}
Existing studies have found that modeling target information (i.g., the victim of the hateful content) can help improve hateful meme classification~\cite{DBLP:conf/mm/LeeCFJC21}. Therefore, we explore the effect of explicitly including the target information in prompts.

The FHM and HarM datasets are annotated with target information. For our experimental design, we change the prompt template: from ``\textit{It was \textbf{[MASK]}.}'' to ``\textit{It was \textbf{[LABEL\_MASK]} targeting at \textbf{[TARGET\_MASK]}.}''. For example, if it is a hateful meme targeting nationality, the template will be ``\textit{It was \textbf{bad} targeting at \textbf{nationality}}.'' If the meme is non-hateful, the \textbf{[TARGET\_MASK]} will be replaced with \textbf{nobody}. During model training, we model the loss from prediction of \textbf{[LABEL\_MASK]} in the inference instance. %The masked word for labels and targets will be provided in demonstrations.
%Similar to our earlier experiments, we perform the target augmentation experiments on full training instances and few-shot setting. 
% For instance, FHM hateful memes are annotated with five target categories: \textit{race}, \textit{disability}, \textit{nationality}, \textit{sex} and \textit{religion}. Hateful memes on HarM are annotated with four target categories: \textit{society}, \textit{individual}, \textit{community} and \textit{organization}.

Table~\ref{tab:targets} shows the results of the \textsf{PromptHate} performance with and without target information. We observe marginal differences in performance after modelling target information in prompts. A possible reason may be that learning to extract targets in memes adds auxiliary burden to the model. To better utilize target information, a more sophisticated strategy may be needed than the current simple approach.

\begin{table}[t]
\small
\centering
  \begin{tabular}{c|cc|cc}
    \hline
    \textbf{Model}& \multicolumn{2}{c|}{\textbf{FHM}} & \multicolumn{2}{c}{\textbf{HarM}}\\
    & AUC. & Acc. & AUC. & Acc.\\
    \hline\hline
    %w/o Target &82.27  &72.36 &90.63 &82.46\\
    %w Target & 82.08&72.28 &88.62 &80.00 \\
    w/o Target &81.45  &72.98 &90.96 &84.47\\
    w Target &81.10 &71.44 & 89.00&82.97 \\
    %\hline
    %\multirow{2}{*}{Few-Shot} & w/o Target &69.30&63.76&83.95&73.53\\
   % & w Target &62.76&56.28&70.44&60.58\\
\hline
\end{tabular}
\caption{\textsf{PromptHate} without and with target.}
  \label{tab:targets}
\end{table}
In Table~\ref{tab:comp-vis}, we visualize \textsf{PromptHate}'s prediction results on sample FHM memes. Incorrect predictions are labelled in red while the pie chart presents the distributions of the predicted target (i.e., \textbf{[TARGET\_MASK]}) per meme. \textsf{PromptHate} with target information is observed to correctly predict the targets in the hateful meme even when it incorrectly classifies the memes (e.g third meme targeting religion). The right-most meme contains a racial slur `Kenyan skidmark' and seems to have been annotated wrongly as non-hateful. Interestingly, \textsf{PromptHate} with target information indicates it as hateful and targeting race.

The target distributions can improve \textsf{PromptHate}'s interpretability. However, the incorrect class prediction also highlights the difficulty of hateful meme classification. The task may require more than target comprehension to achieve good performance.

%Specifically, for the first meme on the left, the model correctly predicted that it was targeting race when there was no mention of the race in the meme.

\begin{table*}[h]
\small
  \centering
  
  \begin{tabular}{|c|c|c|c|c| }
    \hline
    \textbf{Meme} & \begin{minipage}[b]{0.35\columnwidth}
		\centering
		\raisebox{-.5\height}{\includegraphics[width=\linewidth]{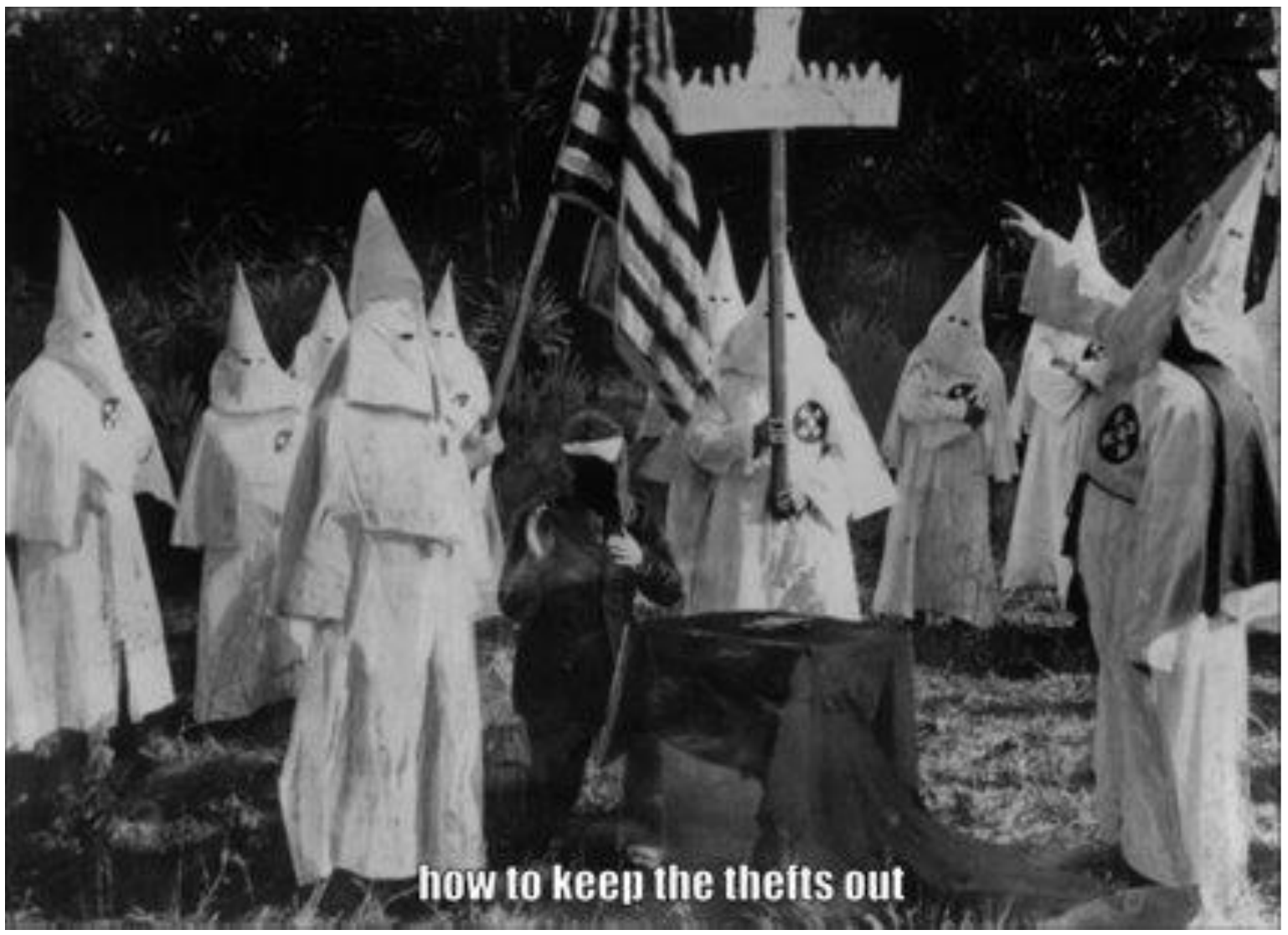}}
	\end{minipage} &
    \begin{minipage}[b]{0.35\columnwidth}
		\centering
		\raisebox{-.5\height}{\includegraphics[width=\linewidth]{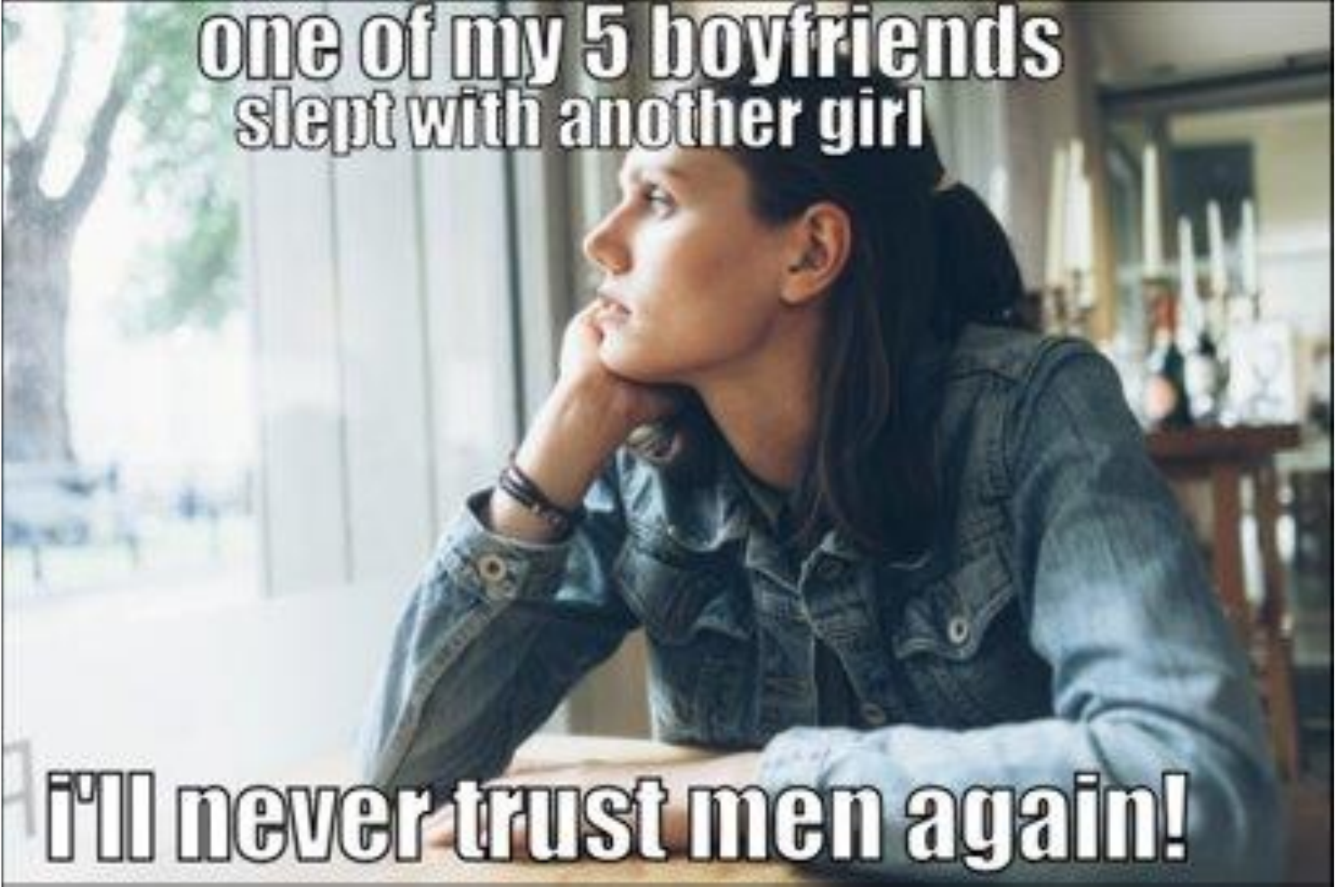}}
	\end{minipage} &
    \begin{minipage}[b]{0.35\columnwidth}
		\centering
		\raisebox{-.5\height}{\includegraphics[width=\linewidth]{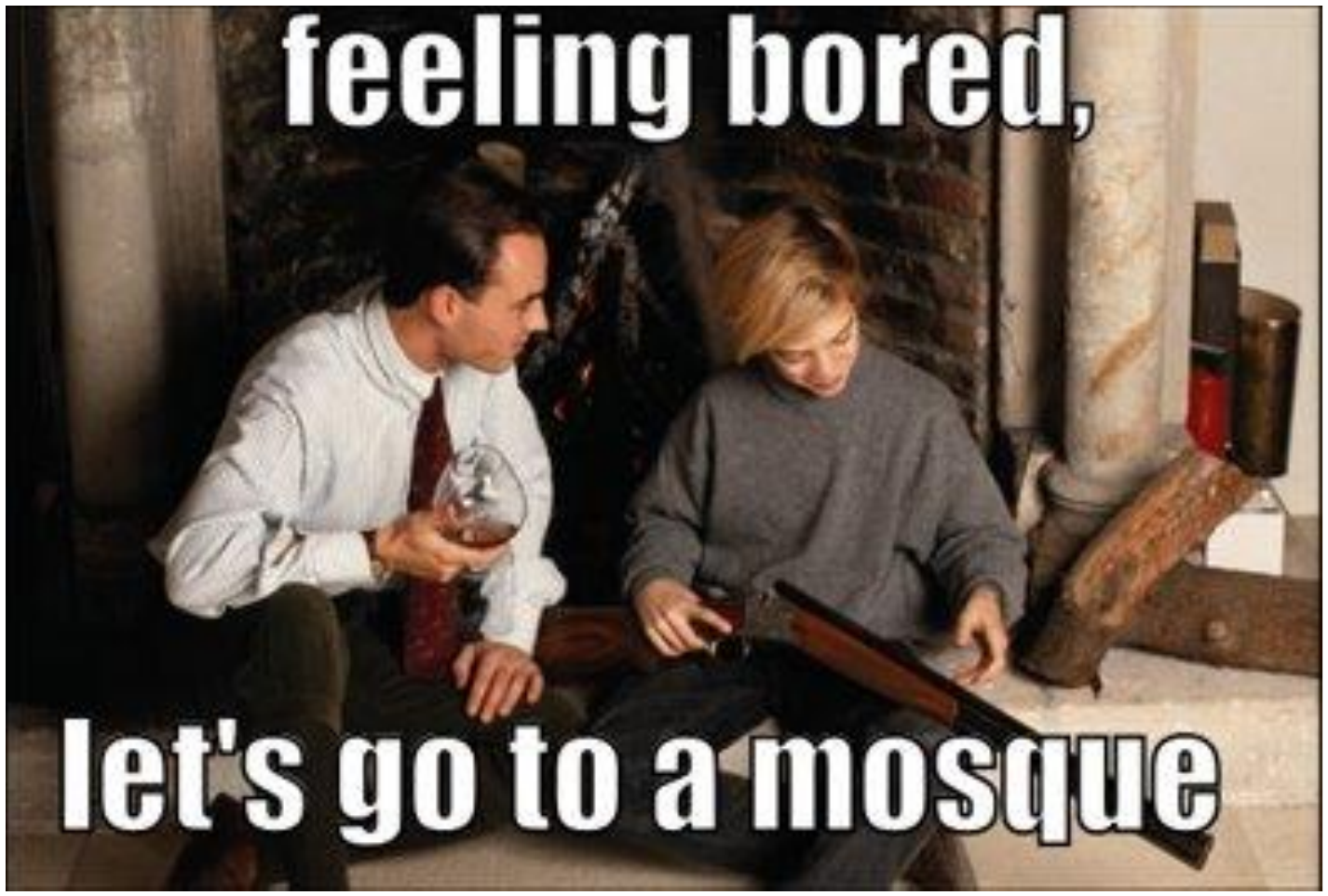}}
	\end{minipage} &
    \begin{minipage}[b]{0.35\columnwidth}
		\centering
		\raisebox{-.5\height}{\includegraphics[width=\linewidth]{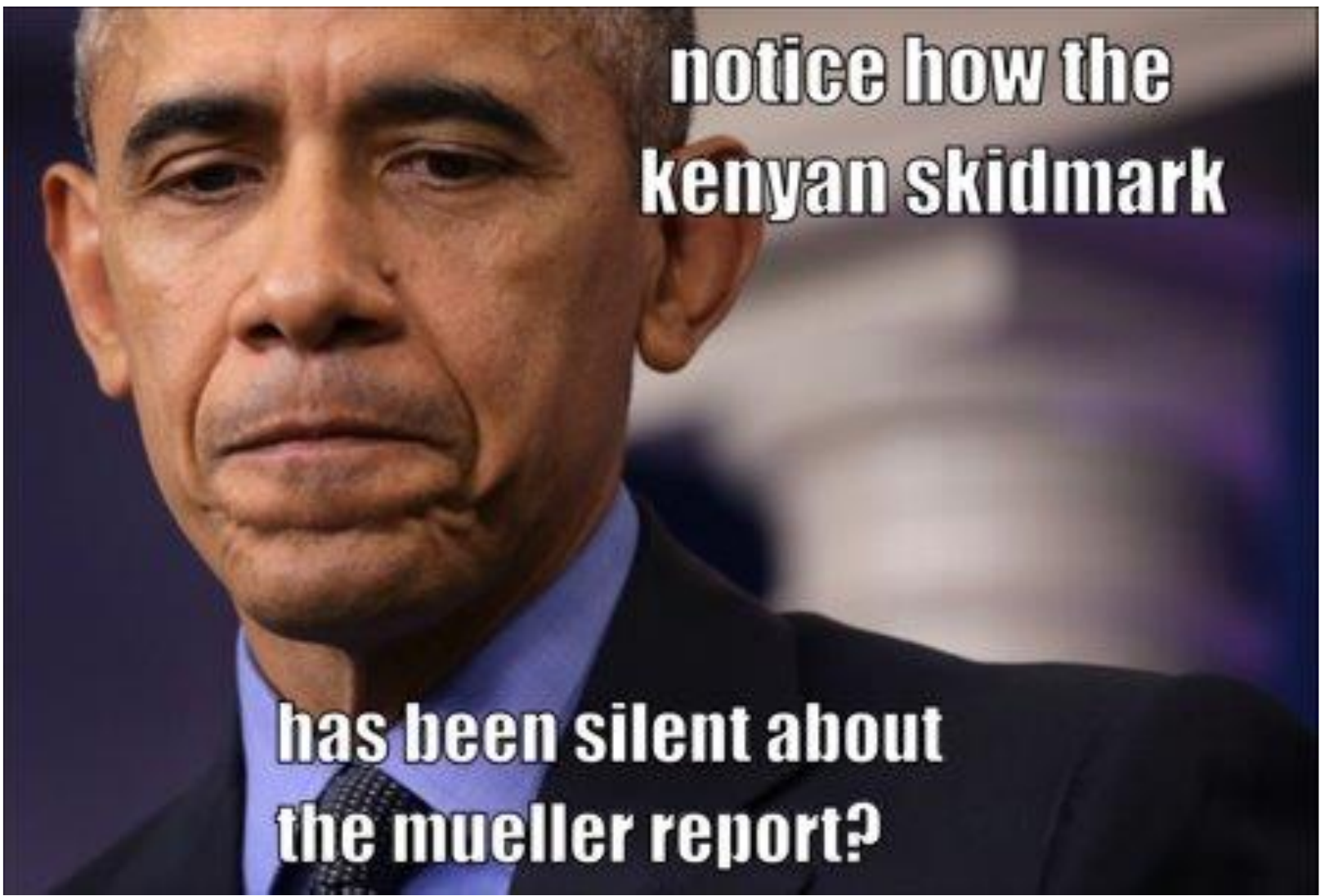}}
	\end{minipage}\\
    \hline
    \textbf{Target Distributions} & \begin{minipage}[b]{0.35\columnwidth}
		\centering
		\raisebox{-.5\height}{\includegraphics[width=\linewidth]{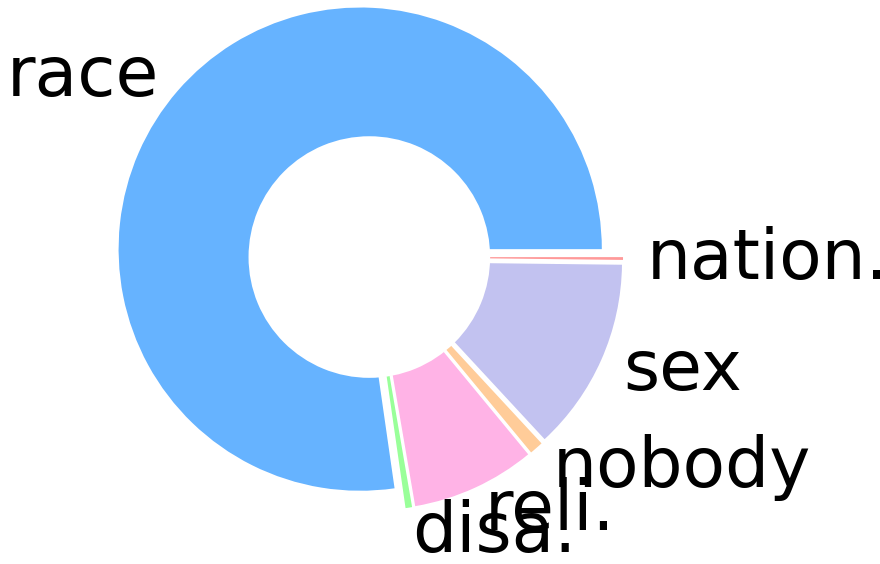}}
	\end{minipage} &
    \begin{minipage}[b]{0.35\columnwidth}
		\centering
		\raisebox{-.5\height}{\includegraphics[width=\linewidth]{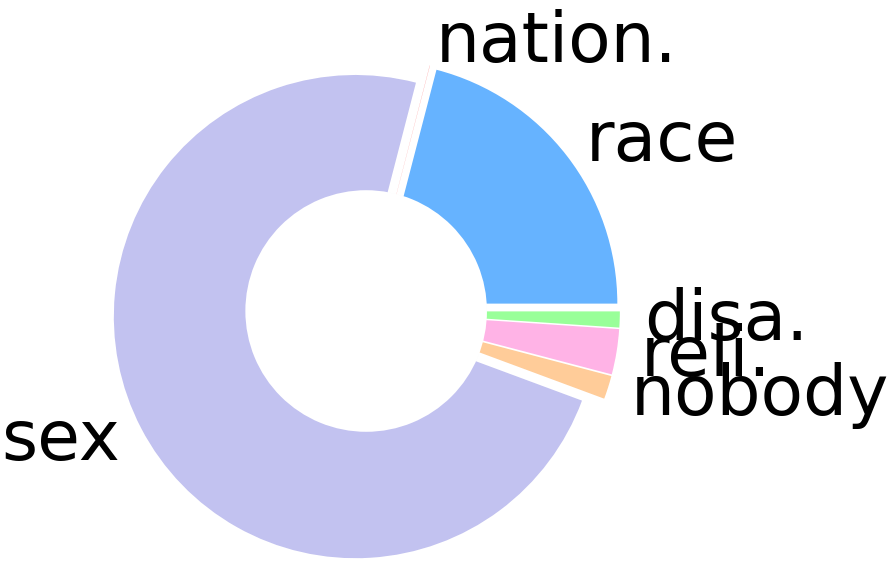}}
	\end{minipage} &
   \begin{minipage}[b]{0.35\columnwidth}
		\centering
		\raisebox{-.5\height}{\includegraphics[width=\linewidth]{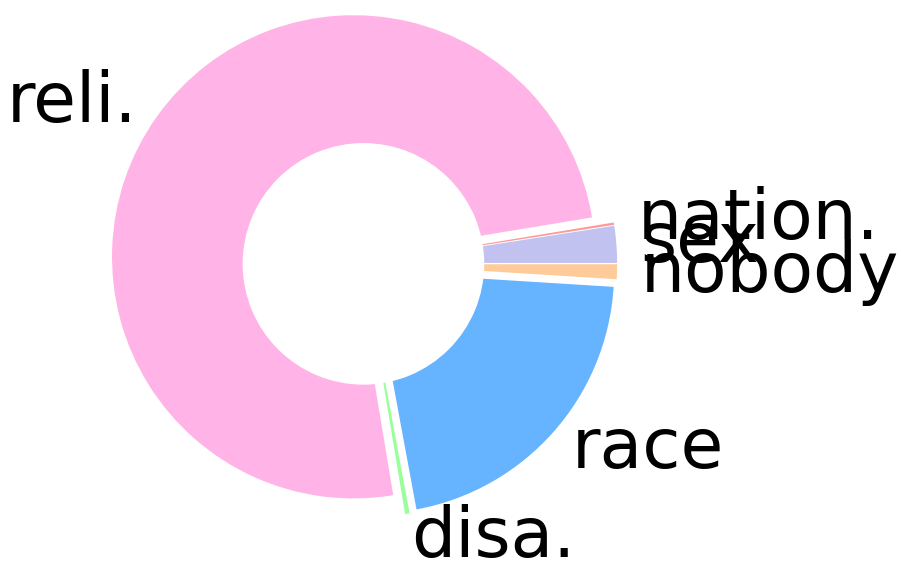}}
	\end{minipage} &
    \begin{minipage}[b]{0.38\columnwidth}
		\centering
		\raisebox{-.5\height}{\includegraphics[width=\linewidth]{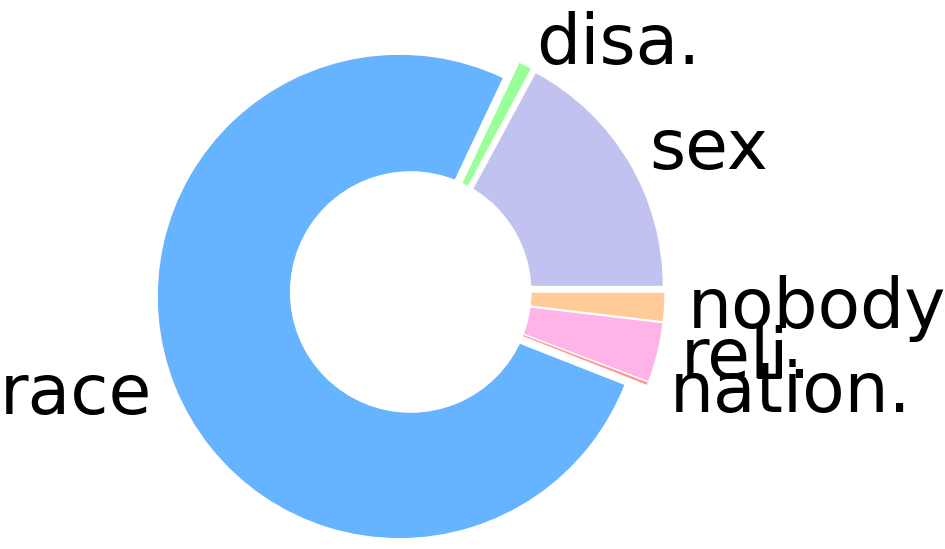}}
	\end{minipage}
   \\\hline
	\textbf{w Target.}  & Hateful & Hateful &\color{red}Non-Hateful  & Hateful \\
	\hline
	\textbf{w/o Target} &\color{red} Non-hateful & \color{red} Non-hateful&  Hateful& Non-hateful \\
	\hline
	\textbf{Ground Truth}  & Hateful (race) & Hateful (sex) & Hateful (religion) & Non-Hateful? \\
	\hline 
    \end{tabular}
    \caption{Example predictions of \textsf{PromptHate} with and without target information. Incorrect prediction in {\color{red} red.} Ground truth for the right-most meme is questionable.}.  
  \label{tab:comp-vis}
\end{table*}

\begin{table*}[h]
\small
  \centering
  %\begin{tabular}{|c|p{3.2cm}|p{3.2cm}|p{3.2cm}|p{3.2cm}| }
  \begin{tabular}{|c|p{4cm}|p{4cm}|p{4cm}| }
    \hline
    \textbf{Meme} & \begin{minipage}[b]{0.5\columnwidth}
		\centering
		\raisebox{-.5\height}{\includegraphics[width=\linewidth]{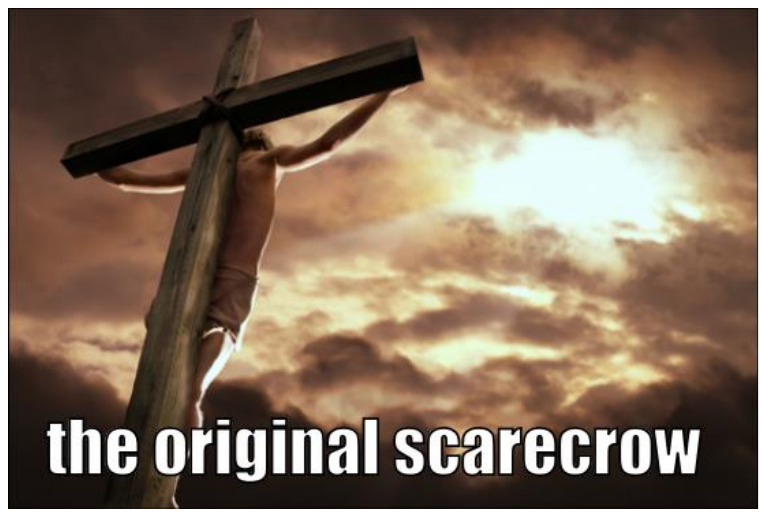}}
	\end{minipage} &
    %\begin{minipage}[b]{0.35\columnwidth}
	%	\centering
	%	\raisebox{-.5\height}{\includegraphics[width=\linewidth]{error/error-terror.pdf}}
	%\end{minipage} &
    \begin{minipage}[b]{0.5\columnwidth}
		\centering
		\raisebox{-.5\height}{\includegraphics[width=\linewidth]{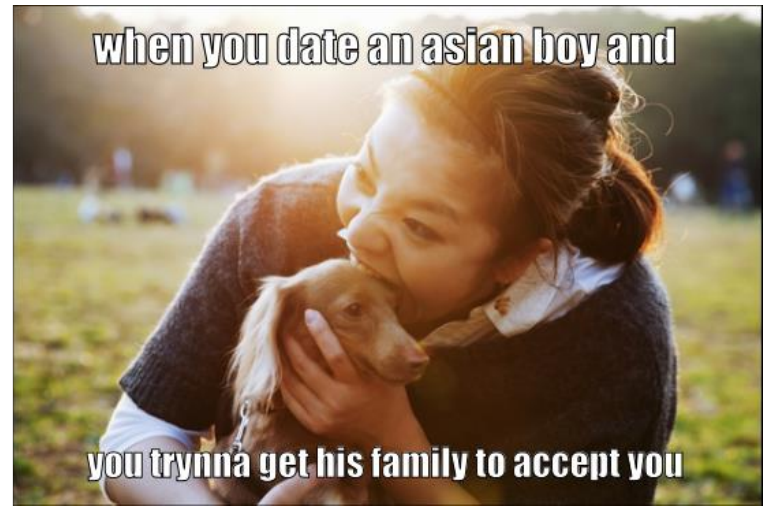}}
	\end{minipage} &
    \begin{minipage}[b]{0.5\columnwidth}
		\centering
		\raisebox{-.5\height}{\includegraphics[width=\linewidth]{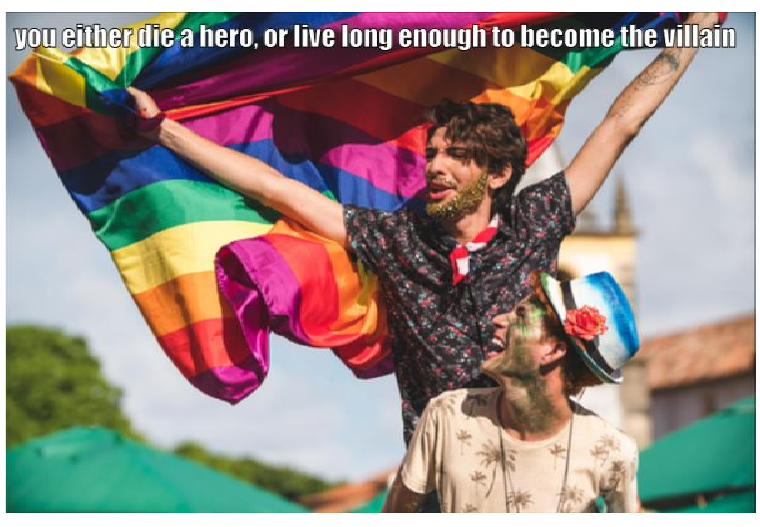}}
	\end{minipage}\\\hline
    %\textbf{Ground Truth} & Hateful & Hateful & Hateful & Non-Hateful\\\hline
    \textbf{Ground Truth}  & Hateful (religion) & Hateful (race) & Non-Hateful\\\hline
    %\textbf{Prediction} & \color{red} Non-hateful & \color{red} Non-hateful & \color{red} Non-hateful & \color{red} Hateful\\\hline
    \textbf{Prediction} & \color{red} Non-hateful & \color{red} Non-hateful & \color{red} Hateful\\\hline
    %\textbf{Meme text} &the original scarecrow &welcome to melbourne &when you date an asian boy and you trynna get his family to accept you&you either die a hero, or live long enough to become the villain\\\hline
    \textbf{Meme text} &the original scarecrow &when you date an asian boy and you trynna get his family to accept you&you either die a hero, or live long enough to become the villain\\\hline
    %\textbf{Captions} &builder crucified on the cross &soldiers stand in front of a building & portrait of a young woman with her pet dog& a man dressed as a rainbow holding a  flag and dancing in the crowd\\ \hline
    \textbf{Captions} &builder crucified on the cross & portrait of a young woman with her pet dog& a man dressed as a rainbow holding a  flag and dancing in the crowd\\ \hline
    
    %\textbf{Entity} & Crucifix Life, Resurrection of Jesus, Spiritual death, Kirkmont Presbyterian Church Salvation, jesus died in the cross& Jihadism, United Nations Editorial Jihadism& None&Rainbow flag, Flag bisexual, Transgender flags, Bisexuality \\\hline
    
    \textbf{Entity} & Crucifix Life, Resurrection of Jesus, Spiritual death, jesus died in the cross & none& Rainbow flag, Flag bisexual, Transgender flags, Bisexuality \\\hline
    
    %\textbf{Demographics} &None & None &Black female & Latino Hispanic Male\\\hline
    \textbf{Demographics} &None &Black female & Latino Hispanic Male\\\hline
    \end{tabular}
    \caption{Error analysis of wrongly predicted memes. Incorrect prediction in {\color{red} red}}
  \label{tab:error}
\end{table*}

\subsection{Error Analysis}
Besides analyzing \textsf{PromptHate}'s quantitative performance, we also examine its classification errors. Table~\ref{tab:error} illustrates three selected \textsf{PromptHate}'s incorrect predictions. 

From the examples, we notice that the captions generally describe the contents of images. However, it may ignore some essential attributes for hateful meme detection. For instance, the captions are unable to capture important information such as ``\textit{Jesus}''. This missing information is supplemented by the augmented image tags (i.e., the entities and demographic of memes). Nevertheless, we also observed that even after augmentation with additional descriptions for the images, \textsf{PromptHate} still makes incorrect predictions for these memes. 

There could be multiple reasons for the incorrect predictions. Firstly, the presented information may still lack adequate context. For instance, in the second meme, the ``biting'' or ``eating'' action is missing from the captions and the addition image description. Thus, \textsf{PromptHate} lacks the context that the meme is ridiculing Asians' ``dog-eating'' behaviour, and is hateful. Secondly, there could be biases learned by the model. For instance, \textsf{PromptHate} may predict the right-most meme to be hateful because of the rainbow flag, an icon for the LGBT community. This icon may be heavily used by memes attacking the LGBT community. Lastly, even more advanced reasoning may be required to understand the hateful context in certain meme. 
In the first case, \textsf{PromptHate} is unable to reason that the meme implies that Jesus is merely an object hung up to scare away birds, which leads to the hatefulness of the meme. The detection of the hateful meme requires deep multi-modal reasoning.

%% file: Conclusion.tex
We have proposed \textsf{PromptHate}, a simple yet effective multimodal prompt-based framework that prompts pre-trained language models for hateful meme classification. Our evaluations on two publicly available datasets have shown \textsf{PromptHate} to outperform  state-of-the-art baselines. We have conducted fine-grained analyses and case studies on various prompt settings and demonstrated the effectiveness of the prompts on hateful meme classification. We have also performed error analysis and discussed some limitations of the \textsf{PromptHate} model. For future work, we will explore strategies for selecting better demonstrations for \textsf{PromptHate} and add in reasoning modules to improve \textsf{PromptHate}'s utilization of the implicit knowledge in the PLMs. %We will also explore methods to better utilize target information in prompt-based models for hateful meme classification. 

%% file: Limitation.tex
We have discussed a number of \textsf{PromptHate}'s limitations in our error analysis (See Section 5.4). Specifically, we noted that in some instances, although \textsf{PromptHate} is able to tap into the implicit knowledge in PLM to improve hateful meme classification, it still lacks the ability to perform advanced reasoning on the contextual information to arrive at correct predictions. We also observe that \textsf{PromptHate} may learn biases from the training data, and debiasing techniques may be required to improve its performance. 

%% file: Appendix.tex
\section{Experiment Settings}
\label{sec:exp-setting}
We train all models using Pytorch on an NVIDIA Tesla V100 GPU, with 32 GB dedicated memory, CUDA-10.2. For pre-trained models (i.e., BERT, RoberTa, VisualBERT), we use the package, \textit{transformers} (version $4.19.2$)  from Huggingface\footnote{https://huggingface.co/}. 
Table~\ref{tab:num-params} lists the parameter count for all models.

\begin{table}[h]
\centering
\begin{small}
  \begin{tabular}{c|c}
    \toprule
    \textbf{Method} & \textbf{\# Params (M)}  \\
    \midrule
    %NMN & 39.43 & 36.99 & 32.41 & 36.14\\
    %NMN (L) & 46.00 & 44.00 & 38.06 & 43.32 \\
    Text BERT & 109.9\\
    Image Region &  1.0\\ 
    \hline
    Late Fusion &110.9 \\ 
    Concat BERT& 111.8\\ 
    MMBT-Region & 111.5  \\
    Visual BERT COCO & 111.8 \\
    ViLBERT CC &252.1 \\
    CLIP BERT  &111.7\\
    MOMENTA &71.9\\
    DisMultiHate &115.6 \\
    \hline
    FT-RoBERTa & 356.4 \\
    PromptHate & 355.4\\
    \midrule
\end{tabular}
\end{small}
\caption{Number of parameters in VQA models.}
  \label{tab:num-params}
\end{table}

The learning rates of models are set empirically. For BERT based models, the learning rate is set to be $2\times10^{-5}$, the same as in~\cite{DBLP:conf/mm/LeeCFJC21}. For RoBERTa-large based models (\textbf{PromptHate} and \textbf{FT-RoBERTa}), following~\cite{DBLP:conf/acl/GaoFC20}, we tested learning rate ranging from $10^{-5}$ to $1.5\times10^{-5}$ and reported the best ones. Specifically, the learning rate is set to be $1.3\times10^{-5}$ and $10^{-5}$ on FHM and HarM datasets, respectively. AdamW is used as the optimizer for all models.
The mini-batch size is set at $16$ during training. As mentioned in Section~\ref{sec:ensemble}, we apply the multi-query ensemble strategy. The number of querying times is set at $4$ on both datasets. It takes one GPU six minutes to train and validate \texttt{PromptHate} per epoch. It takes up $19$ GB dedicated memory for \textsf{PromptHate} training. We use $10$ training epochs for both \textsf{PromptHate} and baselines.

\section{Analysis for Image Captions}
\label{sec:img-cap}
%\textsf{PromptHate} has a number of key components that may influence its performance on the hateful meme classification task. In this section, we perform parametric analysis on two main components in \textsf{PromptHate}: (a) image-to-text caption generation, and (b) multi-query ensemble strategy. 

A key data-preprocessing step in \textsf{PromptHate} is to covert the image into textual captions as input for PLMs. Therefore, the quality and expressiveness of the image captions may affect the prompting and ultimately affect the hateful meme classification performance. To investigate this effect, we experiment with image captions generated with ClipCap~\cite{DBLP:journals/corr/abs-2111-09734} pre-trained on different datasets, namely, MS COCO~\cite{DBLP:conf/eccv/LinMBHPRDZ14,DBLP:journals/corr/ChenFLVGDZ15} and Conceptual Caption (CC)~\cite{DBLP:conf/acl/SoricutDSG18}.  

\begin{table}[t]
\small
\centering
  \begin{tabular}{c|cc|cc}
    \hline
    \textbf{Model}& \multicolumn{2}{c|}{\textbf{FHM}} & \multicolumn{2}{c}{\textbf{HarM}}\\
    & AUC. & Acc. & AUC. & Acc.\\
    \hline\hline
    ClipCap+COCO &78.72&70.20& 87.25&78.38  \\ 
    ClipCap+CC (UC) &80.38&70.08&88.56&81.94 \\ 
    ClipCap+CC &81.45 &72.98 &90.96  &84.47\\
\hline
\end{tabular}
\caption{\textsf{PromptHate} with different image captions.}
  \label{tab:ablation-cap}
\end{table}

\begin{table}[t]
\small
  \centering
  
  \begin{tabular}{|p{1cm}|p{2.4cm}|p{2.4cm}| }
    \hline
    \textbf{Meme} & \begin{minipage}[b]{0.32\columnwidth}
		\centering
		\raisebox{-.5\height}{\includegraphics[width=\linewidth]{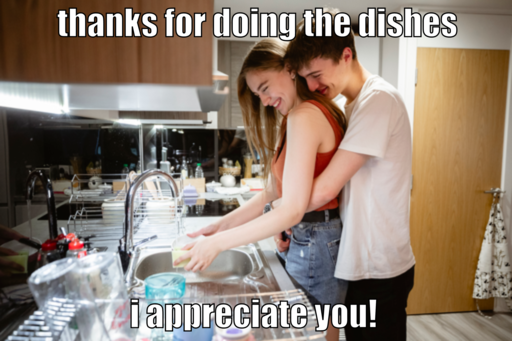}}
	\end{minipage} &
    \begin{minipage}[b]{0.32\columnwidth}
		\centering
		\raisebox{-.5\height}{\includegraphics[width=\linewidth]{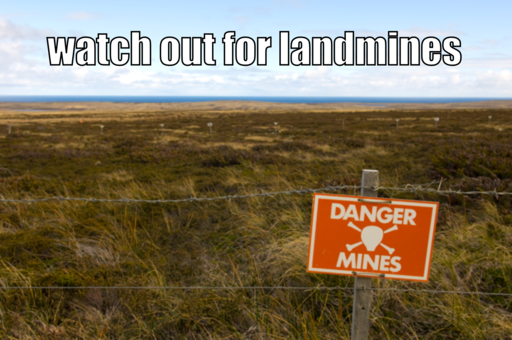}}
	\end{minipage}
	\\ \hline
    \textbf{ClipCap} \textbf{+COCO} &a man and a woman are in a kitchen.
    &a sign that is on the side of a hill.
	\\\hline
	\textbf{ClipCap} \textbf{+CC (UC)} &thank you for the dishes!
    &a warning sign on a hillside.
    \\ \hline
    \textbf{ClipCap} \textbf{+CC} &young couple in love looking at each other in kitchen.
    &warning sign at the entrance to the quarry. 
    \\ \hline
    \end{tabular}
    \caption{Example captions generated for meme images.}
  \label{tab:cap-comp}
\end{table}

Table~\ref{tab:ablation-cap} shows \textsf{PromptHate}'s performance with image captions generated using ClipCap pre-trained on COCO (\textbf{ClipCap+COCO}) and CC (\textbf{ClipCap+CC}). We observe that the ClipCap pre-trained on CC performs better than that pre-trained with COCO. A possible reason could be that the CC dataset is mainly images from web pages and rather more similar to meme images. For instance, considering the examples in Table~\ref{tab:cap-comp}, we notice that ClipCap pre-trained on CC provided more detailed descriptions (e.g., the relation of the man and the woman of the first meme and the characteristic of the sign in the second meme) compared to COCO. On the other hand, we test the generated captions using the uncleaned (i.e., without removing meme texts on images) meme images (\textbf{ClipCap+CC (UC)}). We notice that when trained with Conceptual Captions, ClipCap+CC (UC) is still able to generate some details (i.e., the characteristic of the sign in the second example). It sometimes generates comments rather than captions that describe images. It is because Conceptual Captions are from the web, and some of them are comments on meme images. Without removing texts, models will link the image to meme images and generate comments rather than captions. The difference in their performance is also significant, suggesting the importance of good quality captions in applying prompt-based models for hateful meme classification.

%% file: emnlp2022.bbl
\begin{thebibliography}{45}
\expandafter\ifx\csname natexlab\endcsname\relax\def\natexlab#1{#1}\fi

\bibitem[{Brown et~al.(2020)Brown, Mann, Ryder, Subbiah, Kaplan, Dhariwal,
  Neelakantan, Shyam, Sastry, Askell, Agarwal, Herbert{-}Voss, Krueger,
  Henighan, Child, Ramesh, Ziegler, Wu, Winter, Hesse, Chen, Sigler, Litwin,
  Gray, Chess, Clark, Berner, McCandlish, Radford, Sutskever, and
  Amodei}]{DBLP:conf/nips/BrownMRSKDNSSAA20}
Tom~B. Brown, Benjamin Mann, Nick Ryder, Melanie Subbiah, Jared Kaplan,
  Prafulla Dhariwal, Arvind Neelakantan, Pranav Shyam, Girish Sastry, Amanda
  Askell, Sandhini Agarwal, Ariel Herbert{-}Voss, Gretchen Krueger, Tom
  Henighan, Rewon Child, Aditya Ramesh, Daniel~M. Ziegler, Jeffrey Wu, Clemens
  Winter, Christopher Hesse, Mark Chen, Eric Sigler, Mateusz Litwin, Scott
  Gray, Benjamin Chess, Jack Clark, Christopher Berner, Sam McCandlish, Alec
  Radford, Ilya Sutskever, and Dario Amodei. 2020.
\newblock Language models are few-shot learners.
\newblock In \emph{Advances in Neural Information Processing Systems: Annual
  Conference on Neural Information Processing Systems, NeurIPS}.

\bibitem[{Chen et~al.(2015)Chen, Fang, Lin, Vedantam, Gupta, Doll{\'{a}}r, and
  Zitnick}]{DBLP:journals/corr/ChenFLVGDZ15}
Xinlei Chen, Hao Fang, Tsung{-}Yi Lin, Ramakrishna Vedantam, Saurabh Gupta,
  Piotr Doll{\'{a}}r, and C.~Lawrence Zitnick. 2015.
\newblock \href {http://arxiv.org/abs/1504.00325} {Microsoft {COCO} captions:
  Data collection and evaluation server}.
\newblock \emph{CoRR}.

\bibitem[{Davison et~al.(2019)Davison, Feldman, and
  Rush}]{DBLP:conf/emnlp/DavisonFR19}
Joe Davison, Joshua Feldman, and Alexander~M. Rush. 2019.
\newblock Commonsense knowledge mining from pretrained models.
\newblock In \emph{Proceedings of the Conference on Empirical Methods in
  Natural Language Processing and the International Joint Conference on Natural
  Language Processing, {EMNLP-IJCNLP}}, pages 1173--1178.

\bibitem[{Devlin et~al.(2019)Devlin, Chang, Lee, and
  Toutanova}]{devlin2018bert}
Jacob Devlin, Ming{-}Wei Chang, Kenton Lee, and Kristina Toutanova. 2019.
\newblock {BERT:} pre-training of deep bidirectional transformers for language
  understanding.
\newblock In \emph{Proceedings of the 2019 Conference of the North American
  Chapter of the Association for Computational Linguistics: Human Language
  Technologies, {NAACL-HLT}}, pages 4171--4186.

\bibitem[{Gao et~al.(2021)Gao, Fisch, and Chen}]{DBLP:conf/acl/GaoFC20}
Tianyu Gao, Adam Fisch, and Danqi Chen. 2021.
\newblock Making pre-trained language models better few-shot learners.
\newblock In \emph{Proceedings of the Annual Meeting of the Association for
  Computational Linguistics and the International Joint Conference on Natural
  Language Processing, {ACL/IJCNLP}}, pages 3816--3830.

\bibitem[{Gomez et~al.(2020)Gomez, Gibert, G{\'{o}}mez, and
  Karatzas}]{DBLP:conf/wacv/GomezGGK20}
Raul Gomez, Jaume Gibert, Llu{\'{\i}}s G{\'{o}}mez, and Dimosthenis Karatzas.
  2020.
\newblock Exploring hate speech detection in multimodal publications.
\newblock In \emph{{IEEE} Winter Conference on Applications of Computer Vision,
  {WACV} 2020}, pages 1459--1467. {IEEE}.

\bibitem[{Gui et~al.(2021)Gui, Wang, Huang, Hauptmann, Bisk, and
  Gao}]{DBLP:journals/corr/abs-2112-08614}
Liangke Gui, Borui Wang, Qiuyuan Huang, Alex Hauptmann, Yonatan Bisk, and
  Jianfeng Gao. 2021.
\newblock \href {http://arxiv.org/abs/2112.08614} {{KAT:} {A} knowledge
  augmented transformer for vision-and-language}.
\newblock \emph{CoRR}.

\bibitem[{He et~al.(2016)He, Zhang, Ren, and Sun}]{DBLP:conf/cvpr/HeZRS16}
Kaiming He, Xiangyu Zhang, Shaoqing Ren, and Jian Sun. 2016.
\newblock Deep residual learning for image recognition.
\newblock In \emph{{IEEE} Conference on Computer Vision and Pattern
  Recognition, {CVPR}}, pages 770--778.

\bibitem[{Hee et~al.(2022)Hee, Lee, and Chong}]{hee2022explaining}
Ming~Shan Hee, Roy Ka-Wei Lee, and Wen-Haw Chong. 2022.
\newblock On explaining multimodal hateful meme detection models.
\newblock In \emph{Proceedings of the ACM Web Conference 2022}, pages
  3651--3655.

\bibitem[{K{\"a}rkk{\"a}inen and Joo(2019)}]{karkkainen2019fairface}
Kimmo K{\"a}rkk{\"a}inen and Jungseock Joo. 2019.
\newblock Fairface: Face attribute dataset for balanced race, gender, and age.
\newblock \emph{arXiv preprint arXiv:1908.04913}.

\bibitem[{Kiela et~al.(2019)Kiela, Bhooshan, Firooz, and
  Testuggine}]{DBLP:conf/nips/KielaBFT19}
Douwe Kiela, Suvrat Bhooshan, Hamed Firooz, and Davide Testuggine. 2019.
\newblock Supervised multimodal bitransformers for classifying images and text.
\newblock In \emph{Visually Grounded Interaction and Language (ViGIL), NeurIPS
  Workshop}.

\bibitem[{Kiela et~al.(2020)Kiela, Firooz, Mohan, Goswami, Singh, Ringshia, and
  Testuggine}]{DBLP:conf/nips/KielaFMGSRT20}
Douwe Kiela, Hamed Firooz, Aravind Mohan, Vedanuj Goswami, Amanpreet Singh,
  Pratik Ringshia, and Davide Testuggine. 2020.
\newblock The hateful memes challenge: Detecting hate speech in multimodal
  memes.
\newblock In \emph{Advances in Neural Information Processing Systems,
  {NeurIPS}}.

\bibitem[{Lee et~al.(2021)Lee, Cao, Fan, Jiang, and
  Chong}]{DBLP:conf/mm/LeeCFJC21}
Roy~Ka{-}Wei Lee, Rui Cao, Ziqing Fan, Jing Jiang, and Wen{-}Haw Chong. 2021.
\newblock Disentangling hate in online memes.
\newblock In \emph{{MM} '21: {ACM} Multimedia Conference}, pages 5138--5147.

\bibitem[{Li et~al.(2019)Li, Yatskar, Yin, Hsieh, and Chang}]{li2019visualbert}
Liunian~Harold Li, Mark Yatskar, Da~Yin, Cho-Jui Hsieh, and Kai-Wei Chang.
  2019.
\newblock \href {http://arxiv.org/abs/1908.03557} {Visualbert: A simple and
  performant baseline for vision and language}.
\newblock \emph{CoRR}.

\bibitem[{Lin et~al.(2014)Lin, Maire, Belongie, Hays, Perona, Ramanan,
  Doll{\'{a}}r, and Zitnick}]{DBLP:conf/eccv/LinMBHPRDZ14}
Tsung{-}Yi Lin, Michael Maire, Serge~J. Belongie, James Hays, Pietro Perona,
  Deva Ramanan, Piotr Doll{\'{a}}r, and C.~Lawrence Zitnick. 2014.
\newblock Microsoft {COCO:} common objects in context.
\newblock In \emph{Computer Vision - {ECCV} European Conference}, volume 8693,
  pages 740--755.

\bibitem[{Lippe et~al.(2020)Lippe, Holla, Chandra, Rajamanickam, Antoniou,
  Shutova, and Yannakoudakis}]{lippe2020multimodal}
Phillip Lippe, Nithin Holla, Shantanu Chandra, Santhosh Rajamanickam, Georgios
  Antoniou, Ekaterina Shutova, and Helen Yannakoudakis. 2020.
\newblock A multimodal framework for the detection of hateful memes.
\newblock \emph{arXiv preprint arXiv:2012.12871}.

\bibitem[{Liu et~al.(2019)Liu, Ott, Goyal, Du, Joshi, Chen, Levy, Lewis,
  Zettlemoyer, and Stoyanov}]{DBLP:journals/corr/abs-1907-11692}
Yinhan Liu, Myle Ott, Naman Goyal, Jingfei Du, Mandar Joshi, Danqi Chen, Omer
  Levy, Mike Lewis, Luke Zettlemoyer, and Veselin Stoyanov. 2019.
\newblock \href {http://arxiv.org/abs/1907.11692} {Roberta: {A} robustly
  optimized {BERT} pretraining approach}.

\bibitem[{Lu et~al.(2019)Lu, Batra, Parikh, and Lee}]{lu2019vilbert}
Jiasen Lu, Dhruv Batra, Devi Parikh, and Stefan Lee. 2019.
\newblock \href {http://arxiv.org/abs/1908.02265} {Vilbert: Pretraining
  task-agnostic visiolinguistic representations for vision-and-language tasks}.
\newblock \emph{CoRR}.

\bibitem[{Min et~al.(2022)Min, Lyu, Holtzman, Artetxe, Lewis, Hajishirzi, and
  Zettlemoyer}]{DBLP:journals/corr/abs-2202-12837}
Sewon Min, Xinxi Lyu, Ari Holtzman, Mikel Artetxe, Mike Lewis, Hannaneh
  Hajishirzi, and Luke Zettlemoyer. 2022.
\newblock \href {http://arxiv.org/abs/2202.12837} {Rethinking the role of
  demonstrations: What makes in-context learning work?}
\newblock \emph{CoRR}.

\bibitem[{Mokady et~al.(2021)Mokady, Hertz, and
  Bermano}]{DBLP:journals/corr/abs-2111-09734}
Ron Mokady, Amir Hertz, and Amit~H. Bermano. 2021.
\newblock \href {http://arxiv.org/abs/2111.09734} {Clipcap: {CLIP} prefix for
  image captioning}.
\newblock \emph{CoRR}.

\bibitem[{Muennighoff(2020)}]{DBLP:journals/corr/abs-2012-07788}
Niklas Muennighoff. 2020.
\newblock \href {http://arxiv.org/abs/2012.07788} {Vilio: State-of-the-art
  visio-linguistic models applied to hateful memes}.
\newblock \emph{CoRR}.

\bibitem[{Petroni et~al.(2019)Petroni, Rockt{\"{a}}schel, Riedel, Lewis,
  Bakhtin, Wu, and Miller}]{DBLP:conf/emnlp/PetroniRRLBWM19}
Fabio Petroni, Tim Rockt{\"{a}}schel, Sebastian Riedel, Patrick S.~H. Lewis,
  Anton Bakhtin, Yuxiang Wu, and Alexander~H. Miller. 2019.
\newblock Language models as knowledge bases?
\newblock In \emph{Proceedings of the Conference on Empirical Methods in
  Natural Language Processing and the International Joint Conference on Natural
  Language Processing, {EMNLP-IJCNLP}}, pages 2463--2473.

\bibitem[{Pramanick et~al.(2021{\natexlab{a}})Pramanick, Dimitrov, Mukherjee,
  Sharma, Akhtar, Nakov, and Chakraborty}]{DBLP:conf/acl/PramanickDMSANC21}
Shraman Pramanick, Dimitar Dimitrov, Rituparna Mukherjee, Shivam Sharma,
  Md.~Shad Akhtar, Preslav Nakov, and Tanmoy Chakraborty. 2021{\natexlab{a}}.
\newblock Detecting harmful memes and their targets.
\newblock In \emph{Findings of the Association for Computational Linguistics:
  {ACL/IJCNLP}}, pages 2783--2796.

\bibitem[{Pramanick et~al.(2021{\natexlab{b}})Pramanick, Sharma, Dimitrov,
  Akhtar, Nakov, and Chakraborty}]{DBLP:conf/emnlp/PramanickSDAN021}
Shraman Pramanick, Shivam Sharma, Dimitar Dimitrov, Md.~Shad Akhtar, Preslav
  Nakov, and Tanmoy Chakraborty. 2021{\natexlab{b}}.
\newblock {MOMENTA:} {A} multimodal framework for detecting harmful memes and
  their targets.
\newblock In \emph{Findings of the Association for Computational Linguistics:
  {EMNLP}}, pages 4439--4455.

\bibitem[{Radford et~al.(2021)Radford, Kim, Hallacy, Ramesh, Goh, Agarwal,
  Sastry, Askell, Mishkin, Clark, Krueger, and
  Sutskever}]{DBLP:conf/icml/RadfordKHRGASAM21}
Alec Radford, Jong~Wook Kim, Chris Hallacy, Aditya Ramesh, Gabriel Goh,
  Sandhini Agarwal, Girish Sastry, Amanda Askell, Pamela Mishkin, Jack Clark,
  Gretchen Krueger, and Ilya Sutskever. 2021.
\newblock Learning transferable visual models from natural language
  supervision.
\newblock In \emph{Proceedings of the International Conference on Machine
  Learning, {ICML}}, volume 139, pages 8748--8763.

\bibitem[{Radford et~al.(2019)Radford, Wu, Child, Luan, Amodei, and
  Sutskever}]{radford2019language}
Alec Radford, Jeff Wu, Rewon Child, David Luan, Dario Amodei, and Ilya
  Sutskever. 2019.
\newblock Language models are unsupervised multitask learners.
\newblock \emph{Technical report, OpenAI}.

\bibitem[{Ren et~al.(2016)Ren, He, Girshick, and Sun}]{ren2016faster}
Shaoqing Ren, Kaiming He, Ross Girshick, and Jian Sun. 2016.
\newblock Faster r-cnn: towards real-time object detection with region proposal
  networks.
\newblock \emph{IEEE transactions on pattern analysis and machine
  intelligence}, 39(6):1137--1149.

\bibitem[{Sandulescu(2020)}]{DBLP:journals/corr/abs-2012-13235}
Vlad Sandulescu. 2020.
\newblock \href {http://arxiv.org/abs/2012.13235} {Detecting hateful memes
  using a multimodal deep ensemble}.
\newblock \emph{CoRR}, abs/2012.13235.

\bibitem[{Schick and
  Sch{\"{u}}tze(2021{\natexlab{a}})}]{DBLP:conf/eacl/SchickS21}
Timo Schick and Hinrich Sch{\"{u}}tze. 2021{\natexlab{a}}.
\newblock Exploiting cloze-questions for few-shot text classification and
  natural language inference.
\newblock In \emph{Proceedings of the Conference of the European Chapter of the
  Association for Computational Linguistics: Main Volume, {EACL}}, pages
  255--269.

\bibitem[{Schick and
  Sch{\"{u}}tze(2021{\natexlab{b}})}]{DBLP:conf/naacl/SchickS21}
Timo Schick and Hinrich Sch{\"{u}}tze. 2021{\natexlab{b}}.
\newblock It's not just size that matters: Small language models are also
  few-shot learners.
\newblock In \emph{Proceedings of the Conference of the North American Chapter
  of the Association for Computational Linguistics: Human Language
  Technologies, {NAACL-HLT}}, pages 2339--2352.

\bibitem[{Schwartz et~al.(2017)Schwartz, Sap, Konstas, Zilles, Choi, and
  Smith}]{DBLP:conf/conll/SchwartzSKZCS17}
Roy Schwartz, Maarten Sap, Ioannis Konstas, Leila Zilles, Yejin Choi, and
  Noah~A. Smith. 2017.
\newblock The effect of different writing tasks on linguistic style: {A} case
  study of the {ROC} story cloze task.
\newblock In \emph{Proceedings of the 21st Conference on Computational Natural
  Language Learning}, pages 15--25.

\bibitem[{Sharma et~al.(2018)Sharma, Ding, Goodman, and
  Soricut}]{DBLP:conf/acl/SoricutDSG18}
Piyush Sharma, Nan Ding, Sebastian Goodman, and Radu Soricut. 2018.
\newblock Conceptual captions: {A} cleaned, hypernymed, image alt-text dataset
  for automatic image captioning.
\newblock In \emph{Proceedings of the Annual Meeting of the Association for
  Computational Linguistics, {ACL}}, pages 2556--2565.

\bibitem[{Suryawanshi et~al.(2020)Suryawanshi, Chakravarthi, Arcan, and
  Buitelaar}]{DBLP:conf/acl-trac/SuryawanshiCAB20}
Shardul Suryawanshi, Bharathi~Raja Chakravarthi, Mihael Arcan, and Paul
  Buitelaar. 2020.
\newblock Multimodal meme dataset (multioff) for identifying offensive content
  in image and text.
\newblock In \emph{Proceedings of the Second Workshop on Trolling, Aggression
  and Cyberbullying, TRAC@LREC}, pages 32--41.

\bibitem[{Talmor et~al.(2020)Talmor, Elazar, Goldberg, and
  Berant}]{DBLP:journals/tacl/TalmorEGB20}
Alon Talmor, Yanai Elazar, Yoav Goldberg, and Jonathan Berant. 2020.
\newblock olmpics - on what language model pre-training captures.
\newblock \emph{Trans. Assoc. Comput. Linguistics}, 8:743--758.

\bibitem[{Trinh and Le(2018)}]{DBLP:journals/corr/abs-1806-02847}
Trieu~H. Trinh and Quoc~V. Le. 2018.
\newblock \href {http://arxiv.org/abs/1806.02847} {A simple method for
  commonsense reasoning}.
\newblock \emph{CoRR}.

\bibitem[{Velioglu and Rose(2020)}]{DBLP:journals/corr/abs-2012-12975}
Riza Velioglu and Jewgeni Rose. 2020.
\newblock \href {http://arxiv.org/abs/2012.12975} {Detecting hate speech in
  memes using multimodal deep learning approaches: Prize-winning solution to
  hateful memes challenge}.
\newblock \emph{CoRR}.

\bibitem[{Yang et~al.(2021)Yang, Gan, Wang, Hu, Lu, Liu, and
  Wang}]{DBLP:journals/corr/abs-2109-05014}
Zhengyuan Yang, Zhe Gan, Jianfeng Wang, Xiaowei Hu, Yumao Lu, Zicheng Liu, and
  Lijuan Wang. 2021.
\newblock \href {http://arxiv.org/abs/2109.05014} {An empirical study of
  {GPT-3} for few-shot knowledge-based {VQA}}.
\newblock \emph{CoRR}.

\bibitem[{Yao et~al.(2021)Yao, Zhang, Zhang, Liu, Chua, and
  Sun}]{DBLP:journals/corr/abs-2109-11797}
Yuan Yao, Ao~Zhang, Zhengyan Zhang, Zhiyuan Liu, Tat{-}Seng Chua, and Maosong
  Sun. 2021.
\newblock \href {http://arxiv.org/abs/2109.11797} {{CPT:} colorful prompt
  tuning for pre-trained vision-language models}.
\newblock \emph{CoRR}.

\bibitem[{Zeng et~al.(2022)Zeng, Wong, Welker, Choromanski, Tombari, Purohit,
  Ryoo, Sindhwani, Lee, Vanhoucke, and
  Florence}]{DBLP:journals/corr/abs-2204-00598}
Andy Zeng, Adrian Wong, Stefan Welker, Krzysztof Choromanski, Federico Tombari,
  Aveek Purohit, Michael~S. Ryoo, Vikas Sindhwani, Johnny Lee, Vincent
  Vanhoucke, and Pete Florence. 2022.
\newblock \href {http://arxiv.org/abs/2204.00598} {Socratic models: Composing
  zero-shot multimodal reasoning with language}.
\newblock \emph{CoRR}.

\bibitem[{Zhang et~al.(2020)Zhang, Liu, Li, and
  Zhu}]{DBLP:journals/corr/abs-2012-04977}
Weibo Zhang, Guihua Liu, Zhuohua Li, and Fuqing Zhu. 2020.
\newblock \href {http://arxiv.org/abs/2012.04977} {Hateful memes detection via
  complementary visual and linguistic networks}.
\newblock \emph{CoRR}.

\bibitem[{Zhou et~al.(2021)Zhou, Yang, Loy, and
  Liu}]{DBLP:journals/corr/abs-2109-01134}
Kaiyang Zhou, Jingkang Yang, Chen~Change Loy, and Ziwei Liu. 2021.
\newblock \href {http://arxiv.org/abs/2109.01134} {Learning to prompt for
  vision-language models}.
\newblock \emph{CoRR}.

\bibitem[{Zhou et~al.(2022)Zhou, Yang, Loy, and
  Liu}]{DBLP:journals/corr/abs-2203-05557}
Kaiyang Zhou, Jingkang Yang, Chen~Change Loy, and Ziwei Liu. 2022.
\newblock \href {http://arxiv.org/abs/2203.05557} {Conditional prompt learning
  for vision-language models}.
\newblock \emph{CoRR}.

\bibitem[{Zhou and Chen(2020)}]{zhou2020multimodal}
Yi~Zhou and Zhenhao Chen. 2020.
\newblock Multimodal learning for hateful memes detection.
\newblock \emph{arXiv preprint arXiv:2011.12870}.

\bibitem[{Zhu et~al.(2022)Zhu, Lee, and Chong}]{DBLP:conf/websci/ZhuLC22}
Jiawen Zhu, Roy~Ka{-}Wei Lee, and Wen{-}Haw Chong. 2022.
\newblock Multimodal zero-shot hateful meme detection.
\newblock In \emph{WebSci '22: 14th {ACM} Web Science Conference 2022}, pages
  382--389.

\bibitem[{Zhu(2020)}]{zhu2020enhance}
Ron Zhu. 2020.
\newblock \href {http://arxiv.org/abs/2012.08290} {Enhance multimodal
  transformer with external label and in-domain pretrain: Hateful meme
  challenge winning solution}.
\newblock \emph{CoRR}.

\end{thebibliography}
